\documentclass[journal,twocolumn]{IEEEtran}
\usepackage{epsfig}
\usepackage{graphicx}
\usepackage{amsmath}
\usepackage{amssymb,subfigure,cases}
\usepackage{setspace}

\usepackage{color,hyperref}
\usepackage{color,soul}
\usepackage{algorithm}
\usepackage{algorithmic}
\usepackage{multirow}
\usepackage{url}

\usepackage{ amssymb }

\newlength\savedwidth
\newcommand\whline{\noalign{\global\savedwidth\arrayrulewidth
		\global\arrayrulewidth 1.25pt}%
	\hline
	\noalign{\global\arrayrulewidth\savedwidth}}

\ifCLASSINFOpdf
\else
\fi

\definecolor{darkblue}{rgb}{0.0,0.0,1.0}
\hypersetup{colorlinks,breaklinks,
	linkcolor=darkblue,urlcolor=darkblue,
	anchorcolor=darkblue,citecolor=darkblue}

\begin{document}
	
	\title{Embedding Structured Contour and Location Prior in Siamesed Fully Convolutional Networks for Road Detection}
	
	\author{Qi~Wang,\IEEEmembership{~Senior Member,~IEEE}, Junyu~Gao, and Yuan~Yuan,\IEEEmembership{~Senior Member,~IEEE}
		\thanks{This work was supported by the National Natural Science Foundation of China under Grant 61379094, Fundamental Research Funds for the Central Universities under Grant 3102017AX010, the Open Research Fund of Key Laboratory of Spectral Imaging Technology, Chinese Academy of Sciences.
			
			Qi Wang is with the School of Computer Science, with the Unmanned System Research Institute, and with the Center for OPTical IMagery Analysis and Learning, Northwestern Polytechnical University, Xi'an 710072, China (e-mail: crabwq@gmail.com).		
			
			Junyu Gao and Yuan Yuan are with the School of Computer Science and Center for OPTical IMagery Analysis and Learning,
			Northwestern Polytechnical University, Xi'an 710072, Shaanxi, China (e-mail: gjy3035@gmail.com; y.yuan1.ieee@gmail.com).		 		
			
			\copyright 20XX IEEE. Personal use of this material is permitted. Permission from IEEE must be obtained for all other uses, in any current or future media, including reprinting/republishing this material for advertising or promotional purposes, creating new collective works, for resale or redistribution to servers or lists, or reuse of any copyrighted component of this work in other works.
			
		}
	}
	\markboth{{IEEE} Transactions on Intelligent Transportation Systems}%
	{Shell \MakeLowercase{\textit{et al.}}: Bare Demo of IEEEtran.cls for Journals}
	\maketitle
	
	\begin{abstract}
		Road detection from the perspective of moving vehicles is a challenging issue in autonomous driving. Recently, many deep learning methods spring up for this task because they can extract high-level local features to find road regions from raw RGB data, such as Convolutional Neural Networks (CNN) and Fully Convolutional Networks (FCN). However, how to detect the boundary of road accurately is still an intractable problem. In this paper, we propose a siamesed fully convolutional networks (named as ``s-FCN-loc''), which is able to consider RGB-channel images, semantic contours and location priors simultaneously to segment road region elaborately. To be specific, the s-FCN-loc has two streams to process the original RGB images and contour maps respectively. At the same time, the location prior is directly appended to the siamesed FCN to promote the final detection performance. Our contributions are threefold:
		(1)  An s-FCN-loc is proposed that learns more discriminative features of road boundaries than the original FCN to detect more accurate road regions;
		(2)  Location prior is viewed as a type of feature map and directly appended to the final feature map in s-FCN-loc to promote the detection performance effectively, which is easier than other traditional methods, namely different priors for different inputs (image patches);
		(3)  The convergent speed of training s-FCN-loc model is 30\% faster than the original FCN, because of the guidance of highly structured contours.
		The proposed approach is evaluated on KITTI Road Detection Benchmark and One-Class Road Detection Dataset, and achieves a competitive result with state of the arts.

	\end{abstract}

	\section{INTRODUCTION}
	\label{intro}
	
	Recently, autonomous driving has drawn great attention with the popularity of intelligent vehicles. It is a core component for the intelligent transportation systems (ITS) and aims at avoiding accidents during the driving period. Since most traffic accidents happen on road, it is important to precisely detect the road region. An accurate road detection can not only make the vehicle navigate in the correct way but also prompt the driving system to focus on the specific tasks in the street scene, such as lane detection \cite{revilloud2016new}, vehicle detection \cite{6898836}, pedestrian detection \cite{angelova2015pedestrian} and anomaly detection \cite{7564410}. If the road region can be ensured, other detection tasks will also benefit from it.
	
	Although this problem is named as ``road detection'', it is actually a per-pixel classification task which is a type of semantic segmentation or labeling. In the street scene, road can be viewed as a background object, which is usually occluded by foreground objects (such as vehicles, pedestrians or other obstacles) so that the road surface has no definite shape. Thus, road detection is completely different from other object detection tasks which only need to locate the objects using bounding boxes and it is more challenging than the latter.

	\begin{figure}
		\centering
		\includegraphics[width=.45\textwidth]{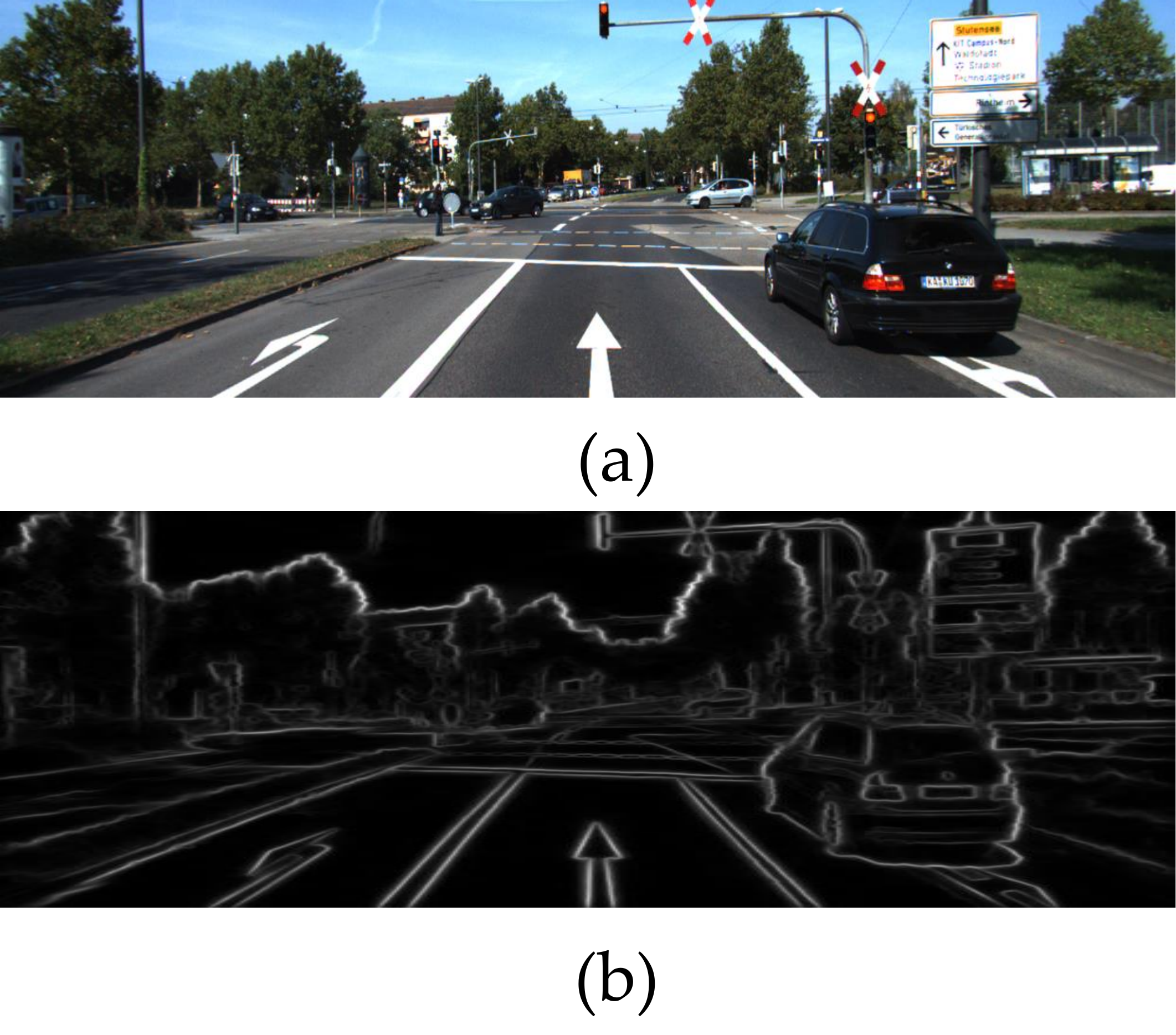}
		\caption{The exemplar display of an original street scene image (a) and the corresponding contour map (b). }\label{Fig-intro}
	\end{figure}

	For detecting road region accurately, some traditional methods (\cite{chen20153d}, \cite{xiao2015crf}, \cite{vitor2014probabilistic} and \cite{7463071}) exploit 3D point clouds or location information by extra sensors such as laser scanner and GPS. In the real world, nevertheless, a human can drive a vehicle safely under the complex traffic environment without the above extra information. Thus, how to dig out deeper vision information is still an important issue, which is our focus in this paper. 
	
	With the rise of deep learning, the Convolutional Neural Networks (CNN) improves the image comprehension by learning more discriminative features (\cite{farabet2013learning}, \cite{DBLP:journals/corr/LongSD14}, \cite{badrinarayanan2015segnet} and \cite{gao2017confroad}). Farabet \emph{et al.} \cite{farabet2013learning} propose a multi-scale CNN to extract dense feature vectors that encode regions of multiple scales centered on each pixel. Fully Convolutional Networks (FCN) \cite{DBLP:journals/corr/LongSD14} is a variant of traditional CNN, which leads to great improvement in many applications, especially in object detection and image semantic segmentation. Seg-net \cite{badrinarayanan2015segnet} proposes an encoder-decoder architecture for image segmentation, in which the encoder is fully convolutional networks and the decoder is deconvolutional networks. The above architectures focus on what an object is but ignore the essential spatial structure and location information in images. In view of this, we introduce spatial structures and location priors into the traditional network:
	
	\subsubsection{\textbf{Spatial Structures}}
	As we all know, the contours in an image represent the essential edges of objects. Different from classic approaches such as  Sobel and
	Canny edge detectors, current methods (e.g. \cite{arbelaez2011contour}, \cite{xiaofeng2012discriminatively} and \cite{dollar2013structured}) focus on detecting semantic edges, which represent the contours of a whole image. Given a contour image of a street scene, human beings are capable of recognizing important objects and their boundaries. Fig. \ref{Fig-intro} illustrates an example of the original image and its corresponding contour map. Based on the above observation, we would like to train a new CNN  model to recognize objects from contour information. 
	
	\subsubsection{\textbf{Location Priors}}
	In the street scenes, the objects' spatial distributions are regular. For example, road region is usually located at the bottom of an image. So how to utilize the location prior is critical to remove the false detection. In the previous methods \cite{brust2015convolutional},  location prior is generated according to the position of each patch in the image, which is cumbersome in the preprocessing stage. Considering these facts, a 2-channel location map is designed to describe the location priors of the whole image and is incorporated in the designed deep model.
	
\begin{figure*}
		\centering
		\includegraphics[width=0.98\textwidth]{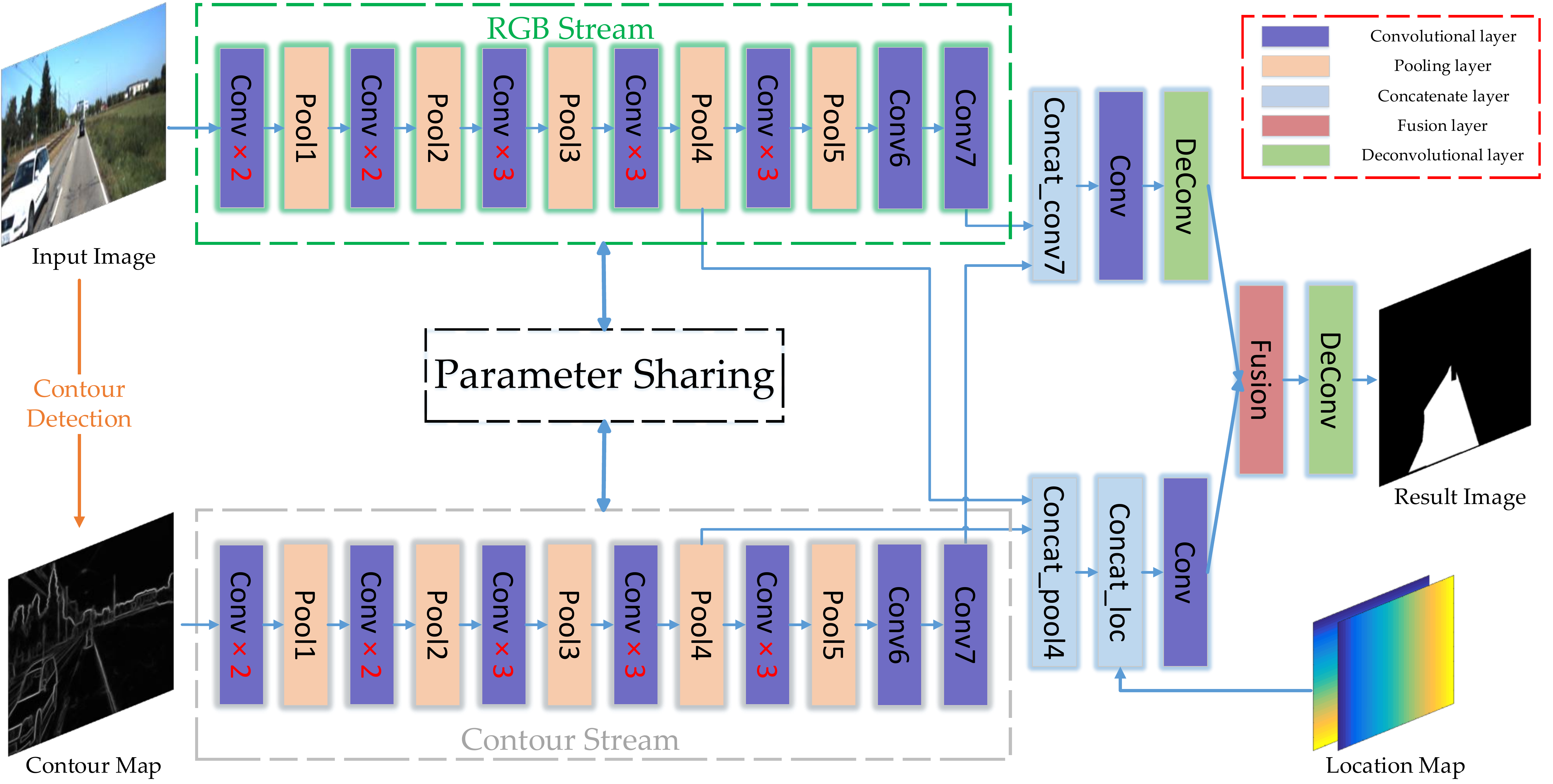}
		\caption{The flowchart of our proposed siamesed FCN with location prior (s-FCN-loc). First, given an input image, the semantic contour map is generated by a fast contour detection. Then, the RGB-channel image and the contour map are fed into s-FCN-loc, which makes s-FCN-loc focus on learning discriminative features of road boundaries and spatial structure in street scene images. At the same time, the location prior is appended to concat\_pool4 layer for alleviating false detection. Finally, the feature map is mapped to each pixel by deconvolution operation to predict the per-pixel road regions.}\label{Fig-overview}
	\end{figure*}

	In summary, the overview of our method is described below. Given an input image, the semantic contour map is firstly generated by Structured Forests (SF) \cite{dollar2013structured}. Then, the RGB image and the contour map are fed into the proposed siamesed fully convolutional networks, exploring  the location feature map simultaneously. Finally, the road region is output by the networks. The concrete flowchart is shown in Fig. \ref{Fig-overview}.
	
	The main contributions of this paper are:
	\begin{enumerate}
		\item[1) ] Propose a siamesed FCN (s-FCN) for road detection, which learns more discriminative features of road boundaries than the original FCN to detect more accurate road regions. The proposed s-FCN consists of two siamesed convoultional streams. It tackles RGB and contour information by sharing the parameters of convolution layers, which prompts the FCN focuses on extracting features of road boundaries. Meanwhile, higher features than raw data also significantly improve the generalization capacity the model.
		\item[2) ] Append the location prior to s-FCN to reduce the mistaken detection. Specifically, the location prior is viewed as a 2-channel feature map to directly concatenate the existing feature map of the network. To our knowledge, the strategy of considering the location prior (s-FCN-loc) is the first time, which is easier than other traditional methods, namely different priors incorporation for different inputs (image patches or superpixels).
		\item[3) ] Accelerate the training process of the original FCN. The convergent speed of training s-FCN-loc is 30\% faster than FCN. The contour maps are regarded as higher-level features than the raw RGB images. The neural network can easily learn more effective semantic representation from the highly structured contour maps, which guides the  model to converge to a good solution more quickly.
	\end{enumerate}
	
	This work is an extension of our earlier conference paper \cite{gao2017confroad}. The more detailed method description and the further experimental analysis, results are shown in this version.
	
	The rest of this paper is organized as follows. Section \ref{related-works} reviews the related work briefly. Section \ref{approach} describes the proposed approach in detail. Section \ref{ex} shows the experimental settings and results on the two challenging datasets and reports the further discussions and analysis about important strategies in our proposed method. Finally, we summarize the work in Section \ref{conclusion}.
	
	\section{RELATED WORK}
	\label{related-works}
	In recent years, many approaches for road detection have been proposed. There are more than $50$ methods on KITTI road detection benchmark since 2013. According to their pipelines, the algorithms usually consist of several important modules: feature extraction, object classification, contextual inference, and priors combination. In this paper, we only briefly review the important works about the  two most related modules: feature extraction and priors combination.
	
	Before the popularity of deep learning, many approaches about road detection are usually comprised of hand-craft feature extraction, per-pixel (superpixel, or block) classification and contextual  refinement.  {\'A}lvarez \emph{et al.} \cite{alvarez2011road} propose illumination invariant features to improve the performance in shadowed street scenes. Mendes \emph{et al.} \cite{mendes2015vision} present a block scheme that classifies small images patches using self-designed features (RGB, grays-cale, entropy, LBP and Leung-Malik filters responses) to efficiently incorporate contextual cues. Since road is background object, which is more cluttered and heterogeneous, Lu \cite{lu2015self} proposes a self-supervised method only using hand-crafted color features without priori knowledge of the road structure. Wang \emph{et al.} \cite{wang2015adaptive} design a novel context-aware descriptor for superpixels by using depth map and transfer labels in a nearest neighbor search set. Yuan \emph{et al.} \cite{yuan2015video} propose an on-line structural learning method for exacting drivable road region from video sequences, which uses the fusion of Dense SIFT, HOG and LBP features for their robustness to intensity change and shadow.
	
	Because of the powerful feature learning ability of CNN, many methods exploit it in recent years. {\'A}lvarez \emph{et al.} \cite{alvarez2012road} train a CNN model from noisy labels to recover the 3D layout of a street image. Then they design a texture descriptor based on a learned color plane fusion to get maximal uniformity in road regions.
	After Long \emph{et al.} \cite{DBLP:journals/corr/LongSD14} proposed FCN, some approaches (such as \cite{laddha2016map}, \cite{DBLP:conf/icra/MendesFW16} and \cite{mohan2014deep}) also adopt this type of architecture to segment road region. Laddha \emph{et al.} \cite{laddha2016map} propose a self-supervised approach which does not require any manual road annotations. Then, they fintune a FCN based on VGG-net \cite{simonyan2014very} using these noisy labels for road detection. Mendes \emph{et al.} \cite{DBLP:conf/icra/MendesFW16} train a FCN model based on Network-in-Network (NiN) architecture, which utilizes large amounts of contextual information. Mohan \cite{mohan2014deep} proposes a deep deconvolutional network in combination with traditional CNNs for feature learning to road detection. Similar to seg-net  architecture \cite{badrinarayanan2015segnet}, Oliveira \emph{et al.} \cite{oliveira2016efficient} propose a smaller network based on an encoder-decoder symmetric network to achieve a near real-time road detection.

	
	Structured priors (such as shape, edge/contour and location) is important to detection results. Many methods also focus on it. He \emph{et al.} \cite{he2004color} model a boundary estimation to improve the detection performance. Yu \emph{et al.} \cite{7299652} proposed a new binary local representation for action recognition from RGB-D video sequences, which adopts an orthogonal projection matrix to preserve the pairwise structure with shape constraints. {\'A}lvarez \emph{et al.} \cite{alvarez2014combining} present an algorithm to estimate road priors by using geographical information systems (GISs), which can provide relevant initial information of the road. Song \emph{et al.} \cite{song2016road} present an algorithm that obtains road boundary information and can be applied to other similar unstructured road environment. Nam \emph{et al.} \cite{nam2015robust} propose a vision-based road detection algorithm, which adopts robust color-based region merging and edge-based filtering mechanisms. Zitnick and Dollar \cite{zitnick2014edge} present a method to locate object proposals based on edge information (the number of edges that are wholly enclosed by a bounding box) in images. Liu \emph{et al.} \cite{liu2015multi} combine CNNs' output and simple edge map via Conditional Random Field for semantic face segmentation. Brust \emph{et al.} \cite{brust2015convolutional} propose convolutional patch networks and incorporate location information into the learning process.

	\section{APPROACH}
	\label{approach}

	In this Section, we first explain the core components of the original FCN \cite{DBLP:journals/corr/LongSD14} in brief. Then the details  of the adopted semantic contour map is described. Next, we show the architecture of the proposed s-FCN. Finally, the strategy of incorporating location priors in s-FCN is explained. 
	
	\subsection{Fully Convolutional Network(FCN)}
	For traditional CNN, the convolutional (``conv'' for short) layers focus on extracting local features in an image, and on the top of multiple conv layers, the fully connected (``fc'' for short) layers integrate those high-level local feature maps into a $n$-D vector by the inner product operation to predict the image's label. Nevertheless, the architecture of this network does not predict the label for each pixel. Until 2015, Long \emph{et al.} \cite{DBLP:journals/corr/LongSD14} propose the Fully Convolutional Networks (FCN) to tackle the dense prediction problem, which replaces all fc layers with conv layers to produce arbitrary-size output. However, since the deep layer's output loses a lot of location and edge clues, the authors of FCN combine deep and shallow layers' feature maps to obtain finer results, which is called as ``FCN-$x$s''. Here $x$ denotes that the fused feature maps need to be $x$ times upsampled to predict the input-image per-pixel label. 
	
	In this paper, we adopt the FCN-16s architecture of VGG16-net \cite{simonyan2014very}, which fuses the pool4 layer and conv7 layer (convolutionalized fc7 of the original network) by a summing operation. It should be noted that pool4's output is cropped and the conv7's is $2$ times upsampled before the fusion for consistent dimensions. VGG16-net can recognize more than $1,000$ categories objects from images, which consists of 13 conv and 3 fc layers. In 2014, it wins the second prize in ImageNet Large Scale Visual Recognition Challenge (ILSVRC) 2014, which achieves 7.32\% Top-5 error in image classification task. 
	
	\subsection{Semantic Contour Map}
	\label{trial}
	
	\begin{figure*}
		\centering
		\includegraphics[width=0.98\textwidth]{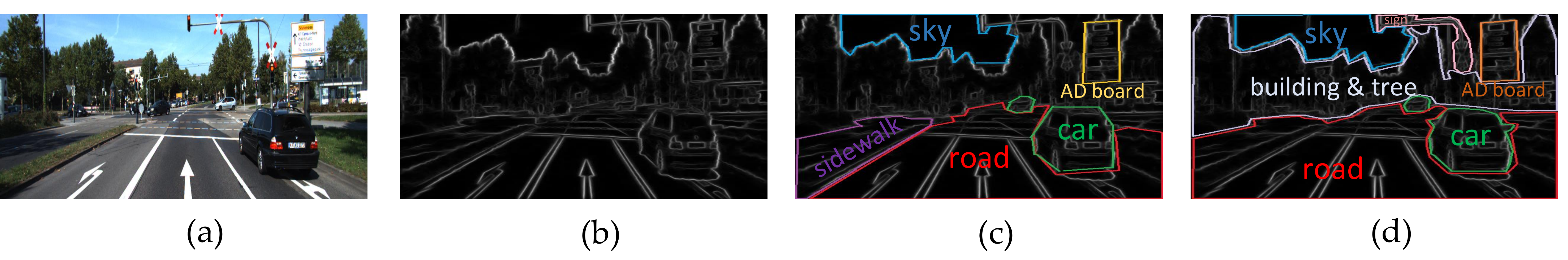}
		\caption{The manually annotated results of the trial in Section \ref{trial}. (a) is the original RGB image; (b) is the contour map, which is given to the subjects; (c) and (d) are the annotated results of two randomly selected subjects. }\label{Fig-contour}
	\end{figure*}
	
	Although FCN fuses the deep and shallow layers' information for alleviating imprecise boundary segmentation to some extent, it is difficult to learn the spatial structure and skeleton clues of an image. Fortunately, semantic contour maps represent them more effectively than traditional edges such as Sobel, Canny, Roberts and so on. In addition, contour map is gray-scale instead of binary image, so that the intensity of contour is quantified. To be specific, Fig. \ref{Fig-contour}(b) is an exemplar of contour map, and the pixels with larger value mean that they are more principal contour in the original image.
	
	In order to validate the point that contour is important for semantic labeling, we design a simple trial that let the examined subjects segment each objects from a semantic contour map of a street scene. And they do not go through any special training to recognize objects from contour images. Fig. \ref{Fig-contour} illustrates the results of this trial. From manual segmentation results, we find human vision are capable of understanding scenes just using the semantic contour map. Although there are some recognition errors contrast to the original image, it is undeniable that the boundary segmentation is elaborate.
	
	The above trail confirms our assumption in a way. Furthermore, we think CNN model can also learn similar ability by supervised training. Therefore, the semantic contour map is generated by SF\footnote{The source code is provided by Piotr Doll{\'a}r in https://github.com/pdollar/edges} \cite{dollar2013structured} and a new stream is added to traditional neural network to process contour information. The concrete description is reported in the next section.
	
	\subsection{Siamesed FCN (s-FCN)}
	
	Our proposed siamesed network is based on FCN-16s \cite{DBLP:journals/corr/LongSD14}, which is shown in Fig. \ref{Fig-overview}. It consists of two streams that handle RGB image and semantic contour map simultaneously. For integrating the two streams' features, the outputs of pool4 and conv7 layer are concatenated together (the sizes are ${\rm{n}} \times 1024 \times 44 \times 44$ and ${\rm{n}} \times 8192 \times 16 \times 16$ respectively, where $n$ denotes the size of each mini-batch). Considering the correspondence of the RGB image and the contour map per-pixel, two streams should interact with each other. Thus, at the training stage, the parameters (kernel weights and biases) of conv layer  of two streams are shared with each other. However, since the contour map is a gray-scale image, its channel number  is not equal to that of RGB image. For sharing parameters, the contour map is replicated on the three channels to be similar to a RGB image.
	
	During the training process, we fine-tune the proposed s-FCN based on the original VGG16-net weights according to the thought of previous section, and minimize the sum of unnormalized soft-max loss for each pixel by SGD.
	
	\subsection{Incorporating Location Priors in s-FCN}
	
	In the street scene, the location prior is important: the objects' spatial distributions are regular. For example, road region is usually located at the bottom of images, and the buildings and trees are on both sides of the road. Thus, utilizing this location prior is essential to remove the false detection. However, the traditional FCN is only sensitive to local appearance features instead of location priors, which causes some unreasonable results. For example, building regions might be mistakenly recognized as road. For alleviating the above problem, \cite{brust2015convolutional} proposes a Convolutional Patch Networks (CPN) with location priors to classify the small patches in images as ``road'' or ``not road''. Specifically, different patches have different location priors, which means that the location prior needs to be generated independently and enter into CPN. However, the CPN's strategy is not flexible in practice. In order to avoid the disadvantage, in this paper, the location prior is viewed as a type of feature map in s-FCN and it can be appended to convolutional layer's output directly. This way, the location feature map is generated only once for all images.

	\begin{figure}
		\centering
		\includegraphics[width=.50\textwidth]{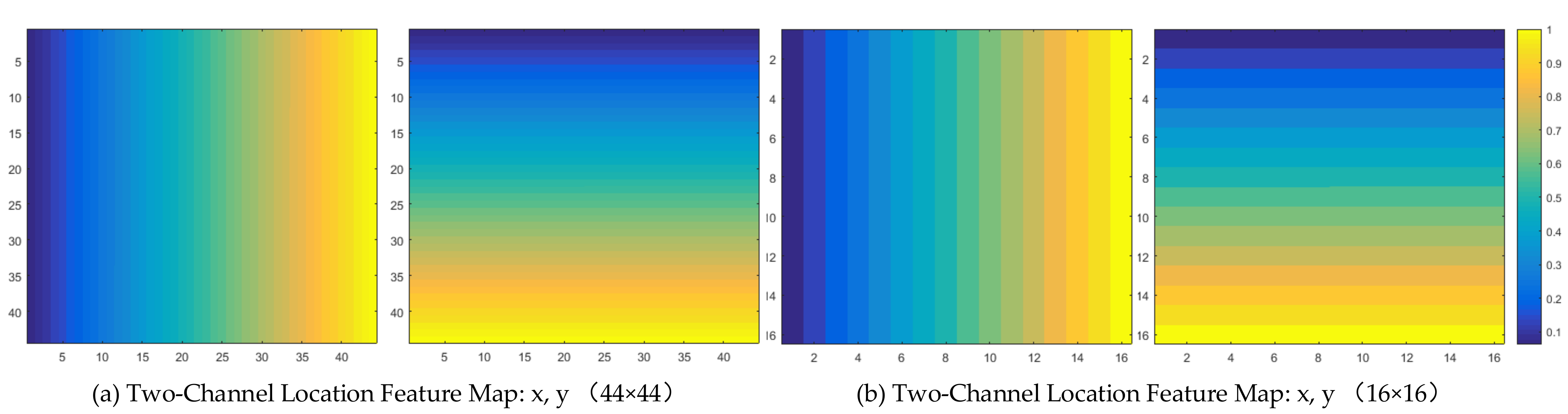}
		\caption{The visualization of the two different sizes of location feature maps. the coordinate values of $x$ and $y$ axis are normalized in $\left[ {0\;,1} \right]$). (a) and (b) illustrate the $44 \times 44$ and $16 \times 16$ location feature maps respectively.}\label{Fig-featuremap}
	\end{figure}
	
	To be specific, location prior is designed as a 2-channel feature map for the $x$ and $y$ axis in the image, which is appended to the last feature map in s-FCN. The value of a position in $x$- or $y$- channel feature map is defined as the coordinate values (normalized in $\left[ {0\;,1} \right]$) of $x$ or $y$ axis in the input image. Since the height and width of the feature map are smaller than the input's, location maps should be resized to the size of the last feature map for concatenating them. It's important to note that there are two final feature maps to be fused in s-FCN: the outputs of concat\_pool4 layer and concat\_conv7 layer (the ${\rm{height}} \times {\rm{width}}$ of the outputs are ${\rm{44}} \times {\rm{44}}$ and ${\rm{16}} \times {\rm{16}}$). Fig. \ref{Fig-featuremap} shows the visualization of the two different sizes of location feature maps. Different colors represent different values: from blue to yellow correspond to $\left[ {0\;,1} \right]$. From Fig. \ref{Fig-featuremap}, the $44 \times 44$ feature map has more accurate location priors than the $16 \times 16$ feature map. Thus, we choose to append the $44 \times 44$ location feature map to concat\_pool4 layer's output. Moreover, we also compare the different effects of the above two strategies by the further experiments in Section \ref{loc-feature-map}.

	\section{EXPERIMENT}
	\label{ex}
	In this section, we respectively report the three comparative results: the original FCN-16s, the proposed s-FCN and the s-FCN-loc on the two challenging road detection datasets. Section \ref{metrics} shows the evaluation criteria in road detection. Section \ref{dataset} briefly describes the two selected dataset. Section \ref{setting} lists some important implementation details and parameter setup in the experiments. Section \ref{KITTI-ex1} and \ref{KITTI-ex2} shows our road detection results in KITTI Dataset and Section \ref{OC-ex} displays our road detection results in OC Dataset. 
	Then, we analyze the convergent speed about different networks architectures in Section \mbox{\ref{speed}}. In Section \mbox{\ref{loc-feature-map}}, we discuss the differences of the different location feature maps. Furthermore, we analyze the generalization capacities of FCN, s-FCN and s-FCN-loc in Section \mbox{\ref{gen}}. Finally, the comparison of Contour Map v.s. Depth Map in s-FCN is in Sectioin \mbox{\ref{com}}.

	\subsection{Metrics}
	\label{metrics}
	For evaluating the algorithm performance, similar to \cite{fritsch2013new}, we adopt the following criteria:
	
	\begin{equation}
	\begin{array}{l}
	{\mathop{\rm Precision}\nolimits}  = \dfrac{{TP}}{{TP + FP}},
	\end{array}
	\end{equation}
	\begin{equation}
	\begin{array}{l}
	{\mathop{\rm Recall}\nolimits}  = \dfrac{{TP}}{{TP + FN}},
	\end{array}
	\end{equation}
	\begin{equation}
	\begin{array}{l}
	Accuracy = \dfrac{{TP + TN}}{{TP + FP + TN + FN}},
	\end{array}
	\end{equation}
	\begin{equation}
	\label{fmeasure}
	\begin{array}{l}
	F{\rm{-}}measure = \left( {1 + {\gamma ^2}} \right)\dfrac{{{\mathop{\rm Precision}\nolimits}  \cdot {\mathop{\rm Recall}\nolimits} }}{{{\gamma ^2}{\mathop{\rm Precision}\nolimits}  + {\mathop{\rm Recall}\nolimits} }},
	\end{array}
	\end{equation}
	where $TP$, $FP$, $TN$ and $FN$ denote the number of true positive, false positive, true negative and false negative samples under a classification threshold $\tau$, respectively. F-measure is a trade off between precision and recall. In this paper, we set $\beta  = 1$ in Eq. \ref{fmeasure} (called ``F1-measure''), which is the harmonic mean of precision and recall. The KITTI benchmark ranks all methods according to max F-measure, which is defined as below:
	\begin{equation}
	\begin{array}{l}
	\max F = \mathop {\arg \max }\limits_\tau  \;F{\rm{-}}measure,
	\end{array}
	\end{equation}
	where $\tau$ is the classification threshold to maximize the F-measure.    
	
	\begin{figure*}
		\centering
		\includegraphics[width=0.98\textwidth]{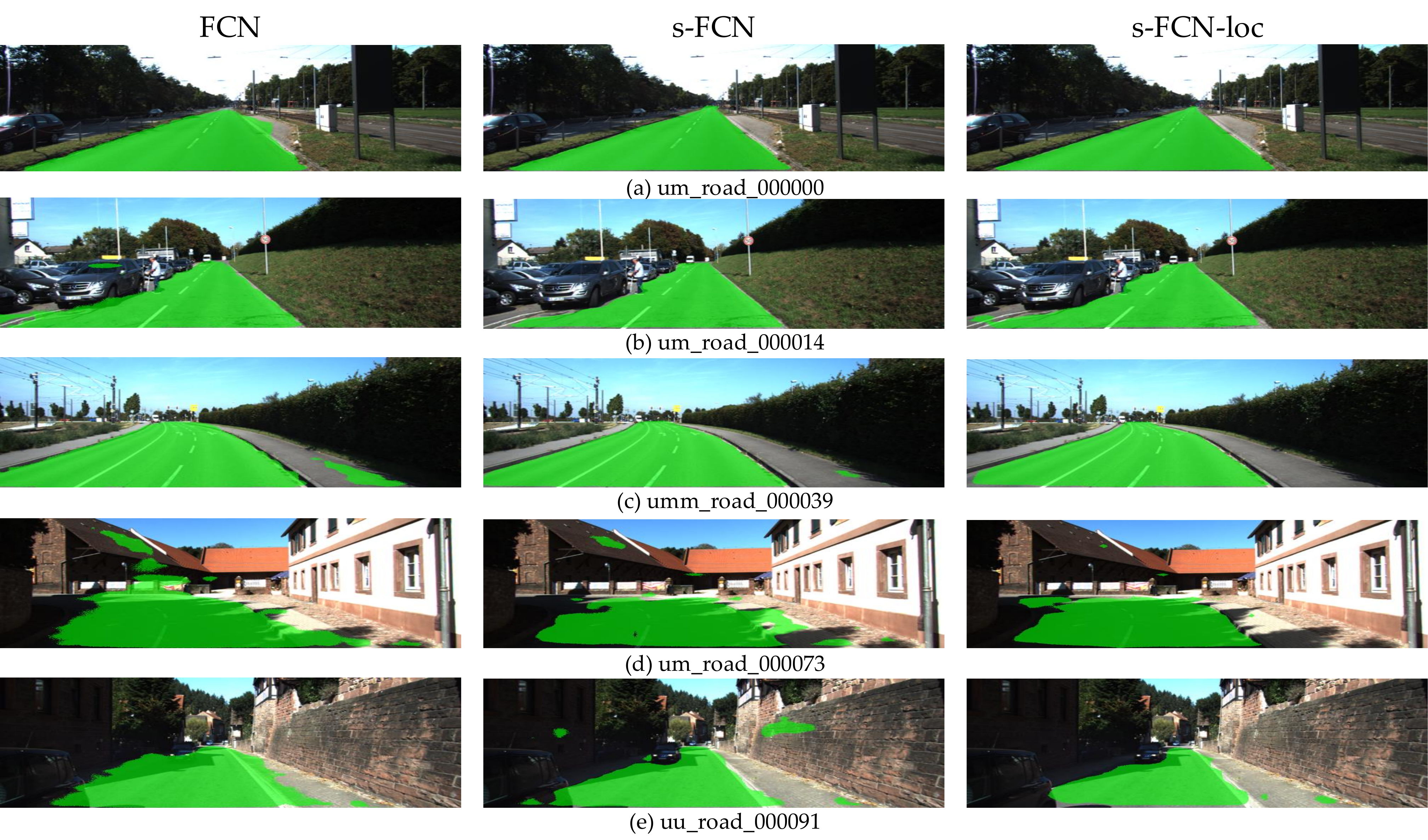}
		\caption{Exemplar results of the different models (from left to right are: the original FCN, s-FCN and s-FCN-loc) on our randomly selected validation set. The green region is the predicted road region.}\label{Fig-val}
	\end{figure*}

	\subsection{Dataset}
	\label{dataset}
	In order to evaluate the proposed approach we select the road detection dataset in KITTI Vision Benchmark Suite \cite{fritsch2013new} \footnote{http://www.cvlibs.net/datasets/kitti/eval\_road.php} (``KITTI dataset'' for short) and One-Class Road Detection Dataset \footnote{http://scrd.josemalvarez.net/} \cite{alvarez2014road} (``OC Dataset'' for short). 
	
	\subsubsection{KITTI Dataset}

	KITTI Dataset consists 579 images (289 training images and 290 testing images respectively) with a resolution of ${\rm{375}} \times {\rm{1242}}$ pixel. The entire data set is divided into three categories, the concrete descriptions of which are shown in Table \ref{tableshow}. The evaluation server of the benchmark ranks all submitted methods according to their max F-measure on the Bird’s Eye View (``BEV'' for short) by assuming a flat real world for the transformation from the perspective image to the BEV space. The benchmark features color stereo images, GPS information and Velodyne laser scans data for each scene. As for this dataset, we only exploit monocular color data to detect road region in the experiment. For showing the effect of each component, the training set is randomly divided into two classes (272 images for training and 17 images for validation).
	
	\begin{table}[htbp]
		\centering
		\caption{The details of KITTI dataset, including the scene category, the numbers of images in training and testing sets.}
		\label{kitti_table}
		\begin{tabular}{c|cc}
			\whline
			Scene category &Training &Testing \\
			\hline
			UU (urban unmarked)   &98  &100  \\
			UM (urban marked two-way)    &95  &96  \\
			UMM (urban marked multi-lane)   &96  &94   \\
			\hline
			URBAN(All)  &289 &290 \\	
			\whline
		\end{tabular}\label{tableshow}
	\end{table}
	
	\subsubsection{OC Dataset}
	The dataset consists of $755$ street scene images $640 \times 480$ pixels. These images include a variety of scenes (e.g. daybreak, morning, noon, afternoon, sunny, cloudy, rainy), which are selected to cover the major challenges in real world. The above challenges contains strong shadows, wet surfaces, sidewalks similar to the road, direct reflections, crowded scenes, lack of lane markings and so on. Because of our supervised machine learning method, the original dataset is randomly divided into two parts (605 images for training and 150 images for testing) to evaluate the proposed algorithms.

	\subsection{Implementation Details and Experimental Setting}
	\label{setting}
	In the entire experiment, original images are resized to  ${\rm{500}} \times {\rm{500}}$ to enter into the neural networks. Contour maps are generated by default parameters (the number of decision trees is 1) of SE-SS in SF \cite{dollar2013structured}. We use two fixed learning rates of ${\rm{1}}{{\rm{0}}^{{\rm{ - 10}}}}$ for weights and ${\rm{2}} \times {\rm{1}}{{\rm{0}}^{{\rm{ - 10}}}}$ for biases, a mini-batch size of 4 images, momentum of $0.99$ and decay of $0.0005$. We also set dropout ratio of 0.5 in conv6 and conv7 layers. Besides, the size of location map is ${\rm{44}} \times {\rm{44}}$ for correspondence of concat\_pool4's output. For the classification threshold $\tau$, we set it as a default value of $0.5$.
	
	For evaluation, the above metrics are computed in the BEV space in KITTI dataset and in the perspective images in OC dataset. 
	
	The experimental environment is equipped with Intel(R) CPU Xeon(R) E5-2697 v2 @ 2.70GHz, 128GB RAM, and four NVIDIA Tesla K80 GPUs. As for the software environment, we modify the standard Caffe\footnote{http://caffe.berkeleyvision.org/} by merging the \#2016\footnote{https://github.com/BVLC/caffe/pull/2016} pull request (PR) of Caffe for saving memory during training process.

	\subsection{KITTI Dataset: Performance on Validation Set}
	\label{KITTI-ex1}
	
	Since the KITTI website only allows the test data to be used strictly for reporting the final results, the stepwise models are evaluated on validation set for showing their effectiveness, and all of the stage results are evaluated on the validation set. Moreover, we also list the result of our full version on the benchmark server to compare with other popular methods in the next subsection. 
	
	Table \ref{table_stage} presents the four metrics (F1-measure, accuracy, precision and recall) of different models on the validation set. Through quantitative results, our proposed full version (``s-FCN-loc'') achieves the best result on the four criteria. In addition, we find each criterion has been improved to some extent except the recall rate, which demonstrates the effectiveness of our proposed siamesed FCN and location prior incorporation.
	
	\begin{table}[htbp]
		\centering
		\caption{Comparison of different stepwise model (the original FCN-16s, s-FCN and s-FCN-loc ) on our selected validation set (in \%).}
		\begin{tabular}{c|cccc}
			\whline
			Methods 				 &F1-measure 		&Acc.			&Pre.		&Rec. 		 		\\
			\hline
			Baseline(FCN)		 	 &92.53			    &97.58				&89.40			&95.90			\\
			s-FCN 		   			 &94.60 		    &98.31 				&93.64 			&95.60			  		\\
			s-FCN-loc 	 &\textbf{95.38} 	&\textbf{98.56} 	&\textbf{94.29}	&\textbf{96.48}			  		\\
			\whline
			
		\end{tabular}\label{table_stage}
	\end{table}

	\begin{figure*}
		\centering
		\includegraphics[width=0.98\textwidth]{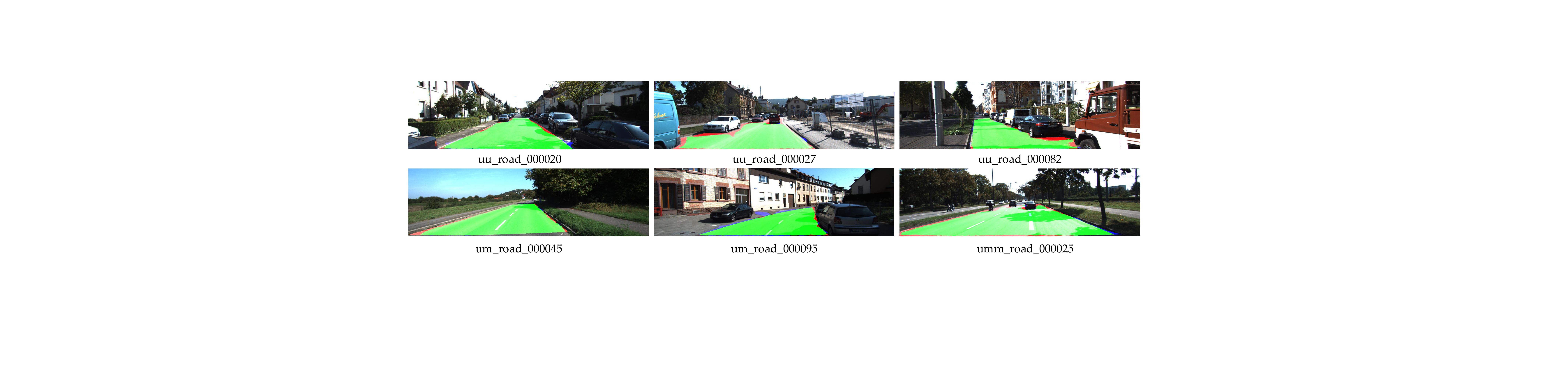}
		\caption{Exemplar results on the KITTI server. The green, blue and red regions denote respectively true positives, false positives and false negatives.}\label{Fig-bench}
	\end{figure*}
	
	\begin{figure*}
		\centering
		\includegraphics[width=1\textwidth]{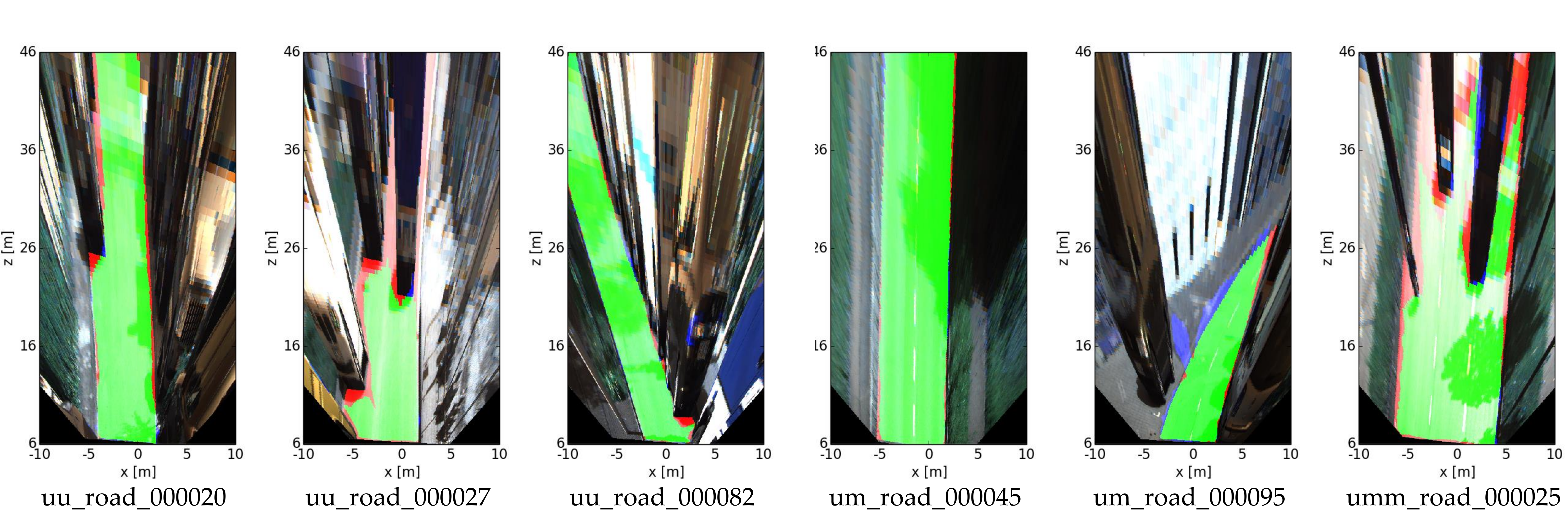}
		\caption{Exemplar BEV space results on the KITTI server. Similarly, the green, blue and red regions denote respectively true positives, false positives and false negatives. A BEV representation covering $\left[ {{\rm{ - }}10m,{\rm{ + }}10m} \right]$ in lateral(x) direction and $\left[ {{\rm{ + }}6m,{\rm{ + }}46m} \right]$ in longitudinal (z) direction is used for evaluation.}\label{Fig-bev-bench}
	\end{figure*}

	In order to analyze the detection performance further and intuitively, Fig. \ref{Fig-val} displays the visualization results of road detection from UM, UMM and UU category. From the first three rows, FCN's results are unclean, especially at the road boundary. For example, the distant sidewalk is mistakenly recognized as road in the third row. By comparison, however, s-FCN and s-FCN-loc are sensitive to the contours of objects, which can segment road region accurately. In the last two sets of exemplars, some building regions are mistaken for road by FCN and s-FCN. The positions of those regions that appear in images are rarely where the road locates. As we can see from the results of the third column, s-FCN-loc incorporating location priors alleviates this problem. These results give a hint that the proposed s-FCN and s-FCN-loc are more effective than the original FCN.

	\subsection{KITTI Dataset: Performance on the Benchmark}
	\label{KITTI-ex2}

	For comparing our proposed s-FCN-loc with other popular methods, we submitted the final results to the KITTI server. Note that the server evaluates the algorithm performance in the BEV space. Table \ref{table_all} shows the results of the first ten real-name submissions \footnote{The leaderboard on the KITTI server includes some anonymous submissions. As for these anonymous submissions, because without their detail information, we do not list them in this paper. It is noted that the proposed model obtains the 8-th prize in all 52 submissions.} and ours in the Urban Road category. Our method achieves a competitive (the second place) result of Max F-measure 93.26\%, which does not differ much from the best 93.43\% of DDN \cite{mohan2014deep}. In the listed methods, DDN \cite{mohan2014deep}, FTP \cite{laddha2016map}, FCN\_LC \cite{DBLP:conf/icra/MendesFW16}, StixelNet \cite{levi2015stixelnet} and MAP \cite{laddha2016map} are deep learning methods and only take advantage of RGB information; NNP \cite{chen20153d}, FusedCRF \cite{xiao2015crf} and ProbBoost \cite{vitor2014probabilistic} exploit 3D information such as stereo vision and LIDAR data; HIM \cite{munoz2010stacked} and CB \cite{mendes2015vision} make use of hand-crafted features to detect road region. In addition to the max F-measure, the results of other four criteria are in the top three. As for runtime, the proposed method is the 4-th place in all $11$ algorithms. Compared with the faster methods, the proposed method is superior to them according to the max F-measure. In general, our method is more competitive than other mainstream algorithms in terms of the detection accuracy and the time performance.  
	
	\begin{table*}[htbp]
		\centering
		\caption{Leaderboard of the Top-10 real-name algorithms on the Urban Road category on the KITTI Vision Benchmark Suite server (in \%). The input source that corresponds to the  $\checkmark$ is exploited by the algorithms. The {\color{red}{red}}, {\color{blue}{blue}} and {\color{green}{green}} fonts respectively represent the {\color{red}{first}}, {\color{blue}{second}} and {\color{green}{third}} place in the corresponding column. }
		
		\begin{tabular}{|c|ccc|ccccc|c|}
			\whline
			\multirow{2}{*}{Methods} &\multicolumn{3}{|c|}{Input Sources} & \multicolumn{5}{|c|}{Metrics} &\multirow{2}{*}{Runtime} \\
			\cline{2-9} 
			&RGB &Stereo &Laser &Max F-measure 	&Precision			&Recall			&FPR 			&FNR		& 		\\
			\whline
			DDN \cite{mohan2014deep}		 			 &\checkmark&		&	&\color{red}{93.43}&\color{red}{95.09}	&91.82			&\color{red}{2.61}	&8.18		&2s  			\\
			\textbf{Ours: s-FCN-loc}			 		 &\checkmark&		&	&\color{blue}{93.26}&\color{blue}{94.16}&\color{green}{92.39}&\color{blue}{3.16}&\color{green}{7.61}&0.4s  		\\
			FTP \cite{laddha2016map} 		   			 &\checkmark&		&	&\color{green}{91.61}&91.04 			&92.20 			&5.00			&7.80		&\color{blue}{0.28s}	\\
			FCN\_LC \cite{DBLP:conf/icra/MendesFW16} 	 &\checkmark&		&	&90.64 		    &90.87 				&90.72			&5.02			&9.28		&\color{red}{0.03s}	\\
			HIM \cite{munoz2010stacked} 		 		 &\checkmark&	&		&90.07			    &\color{green}{91.62}		&89.68	&\color{green}{4.52}	&10.32		&7s  			\\
			NNP \cite{chen20153d} 						 &\checkmark&\checkmark	&&89.68			    &89.67				&89.68 			&5.69			&10.32		&5s  			\\
			StixelNet \cite{levi2015stixelnet}			 &\checkmark&	&		&89.12			    &85.80				&\color{blue}{92.71}&8.45		&\color{blue}{7.29}&1s			\\
			CB \cite{mendes2015vision}					 &\checkmark	&	&	&88.97			    &89.50				&88.44			&5.71			&11.56		&2s  			\\
			FusedCRF \cite{xiao2015crf}					 &\checkmark&	&\checkmark&88.25			    &83.62				&\color{red}{93.44}	&10.08		&\color{red}{6.56}&2s  			\\
			MAP \cite{laddha2016map}		 			 &\checkmark&	&		&87.80			    &86.01				&89.66			&8.04			&10.34		&\color{blue}{0.28s}	\\
			ProbBoost \cite{vitor2014probabilistic}		 &\checkmark&\checkmark&	&87.78			    &86.59				&89.01			&7.60			&10.99		&150s  			\\	
			\whline
			
		\end{tabular}\label{table_all}
	\end{table*}
	
	Fig. \ref{Fig-bench} shows our final results on the KITTI benchmark server. The green, blue and red regions denote respectively true positives, false positives and false negatives. As we can see from the displayed exemplars,  our proposed ``s-FCN-loc'' model has strong  generalization ability from the training set to the testing set. From the ``uu\_road\_000027'' and ``uu\_road\_000082'', however, we find our model still cannot segment accurately in the small corner of road region. The main : deep neural networks output small-size feature map, the receptive field of which is too large to describe the independent features for the small regions. 
	
	Fig. \mbox{\ref{Fig-bev-bench}} shows our BEV space results on the KITTI benchmark server. The $400 \times 800$px BEV image represents the $20m \times 46m$ (meters) real world. From the result of the original ``umm\_road\_000025'' image (in Fig. \ref{Fig-bench}), the performance of our method is good, and only the distant road region can not be segmented accurately. However, in the BEV space, the drawback is magnified. The same phenomenon also exists in other images, such as the ``uu\_road\_000027'' and ``um\_road\_000095''.
	
	Like the proposed s-FCN-loc, FTP \cite{laddha2016map} and FCN\_LC \cite{DBLP:conf/icra/MendesFW16} also belong to Fully Convolutional Networks. But they process small image patches. FTP adopts the traditional FCN provided by DeepLab. FCN\_LC designs a small FCN model based on Network-in-Network (NiN) architecture and takes advantage of large amounts of contextual information. Compared with them, the proposed s-FCN-loc obtains the best on all four criteria according to Table \ref{table_all}. As for visualization results, s-FCN-loc can more accurately segment the road boundary than FTP and FCN\_LC. In addition, the mistaken detection are reduced in s-FCN-LC, which is caused by without location priors in the results of FTP and FCN\_LC.

	\subsection{OC Dataset: Performance}
	\label{OC-ex}
	
	\begin{table}[htbp]
		\centering
		\caption{Comparison of different approaches on OC testing set (in \%). The bold fonts represent the best in the corresponding column.}	
		\begin{tabular}{c|cccc}
			\whline
			Methods 				 						&F1-measure 		&Acc.			&Pre.		&Rec. 		 	\\
			\hline
			FCN-32s \cite{DBLP:journals/corr/LongSD14}   	&94.95			    &96.07			&94.68		&95.22           \\
			FCN-16s	\cite{DBLP:journals/corr/LongSD14}	 	&96.97			    &97.65			&96.66		&97.30			\\
			FCN-8s  \cite{DBLP:journals/corr/LongSD14}      &97.31			    &97.89			&96.56		&98.06           \\
			Seg-net \cite{badrinarayanan2015segnet}			&96.56			    &97.35			&96.86		&96.27		\\
			ENet \cite{paszke2016enet}						&96.21			    &97.02			&94.85		&97.61		\\
			\hline
			\textbf{Our Methods:} \\
			s-FCN 		   			 						&97.46 		    	&98.02 			&97.02 		&\textbf{97.91}\\
			s-FCN-loc 	 									&\textbf{97.56} 	&\textbf{98.10} &\textbf{97.24}	&97.88	\\
			\whline
			
		\end{tabular}\label{table_OC}
	\end{table}
	
	The results of FCN-32/16/8s \cite{DBLP:journals/corr/LongSD14}, Seg-net \cite{badrinarayanan2015segnet}, ENet \cite{paszke2016enet} and our models are listed in Table \ref{table_OC}. FCN-32/16/8s \cite{DBLP:journals/corr/LongSD14} are Fully Convolutional Networks, and the last two models combine deep and shallow layers' feature maps to obtain finer results; Seg-net \cite{badrinarayanan2015segnet} consists of encoder (a Fully Convolutional Networks) and decoder (a DeConvolutional Networks) architecture to predict pixel-wise label. Like Seg-net \cite{badrinarayanan2015segnet}, ENet \cite{paszke2016enet} is also a symmetric encoder-decoder network, which has a smaller architecture than Seg-net \cite{badrinarayanan2015segnet}. Among these algorithms, FCN-32/16/8s \mbox{\cite{DBLP:journals/corr/LongSD14}} and Seg-net \mbox{\cite{badrinarayanan2015segnet}} adopt the VGG-16 net \mbox{\cite{simonyan2014very}} as a pre-trained model. From the table, we can see our full version (``s-FCN-loc'') achieves the best result on the first three criteria. For another criterion - recall, the best performance belongs to ``s-FCN''. It can be seen, after incorporating location priors in s-FCN, the false detection is reduced but the missing detection is increased. In addition, from the last rows, the improvement of incorporating location priors is not significant on OC dataset than KITTI dataset. The main reason is: the distribution area of road in OC dataset is so variable that some infrequent road regions are wrongly removed.
	
\begin{figure*}
		\centering
		\includegraphics[width=0.98\textwidth]{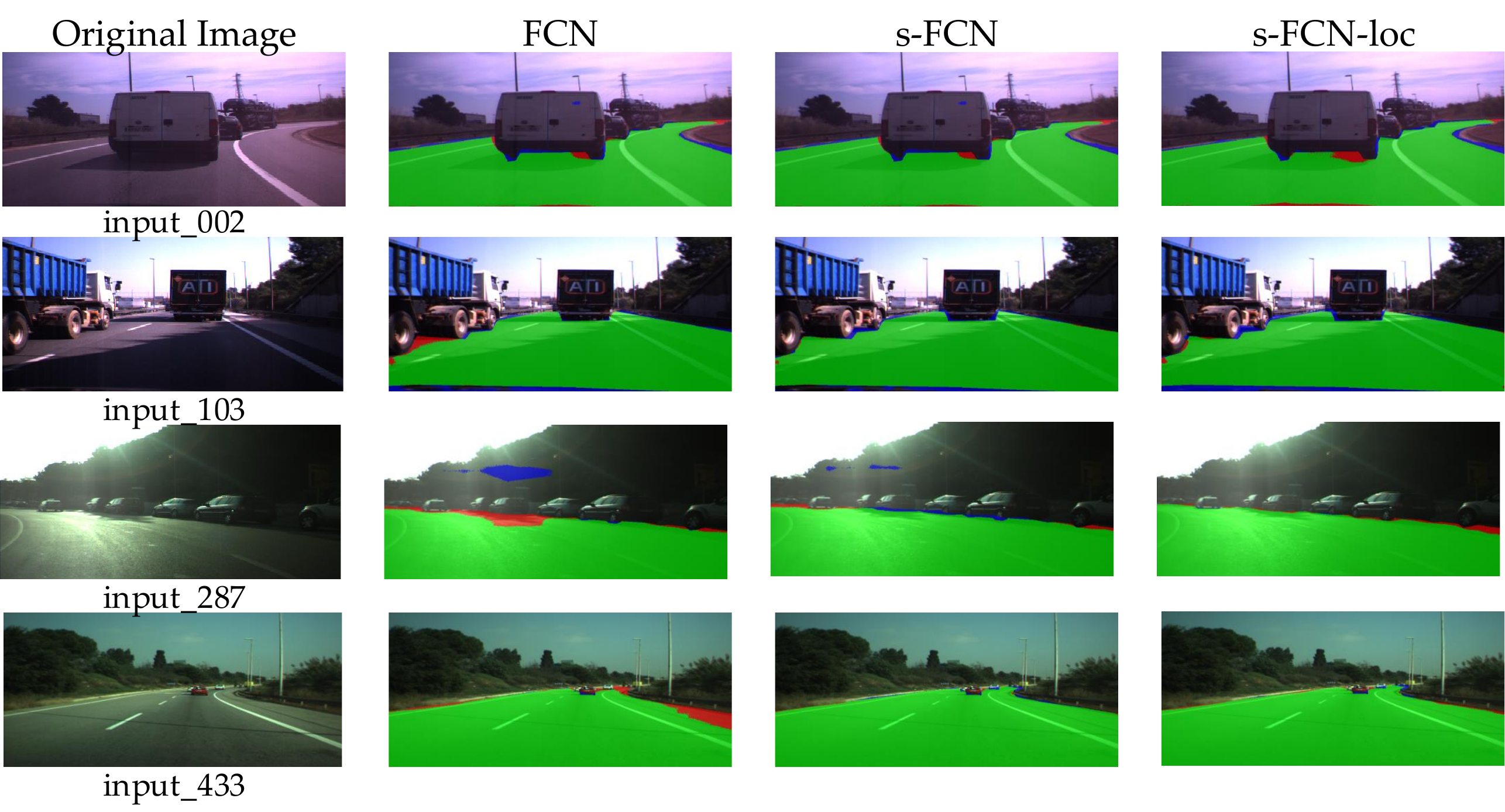}
		\caption{Exemplar results on OC testing set. The green, blue and red regions denote respectively true positives, false positives and false negatives.}\label{Fig-OC}
	\end{figure*}

	Four typical visual results are shown in Fig. \ref{Fig-OC} to intuitively explain the effects of the siamesed FCN and incorporating location priors. From the ``input\_103'' and ``input\_433'', the original FCN cannot segment the road boundaries accurately, but the s-FCN and s-FCN-loc alleviate this problems effectively. In the results of ``input\_002'' and``input\_287'', the original FCN and s-FCN take some inconceivable regions for road (the positions of these regions are where the road is rarely located). After incorporating location priors in s-FCN, the phenomenon is greatly alleviated.

	\subsection{Analysis of Convergent Speed}
	\label{speed}
\begin{figure}
		\centering
		\subfigure[KITTI dataset.] { \label{fig_coner_1} 
			\includegraphics[width=0.46 \columnwidth]{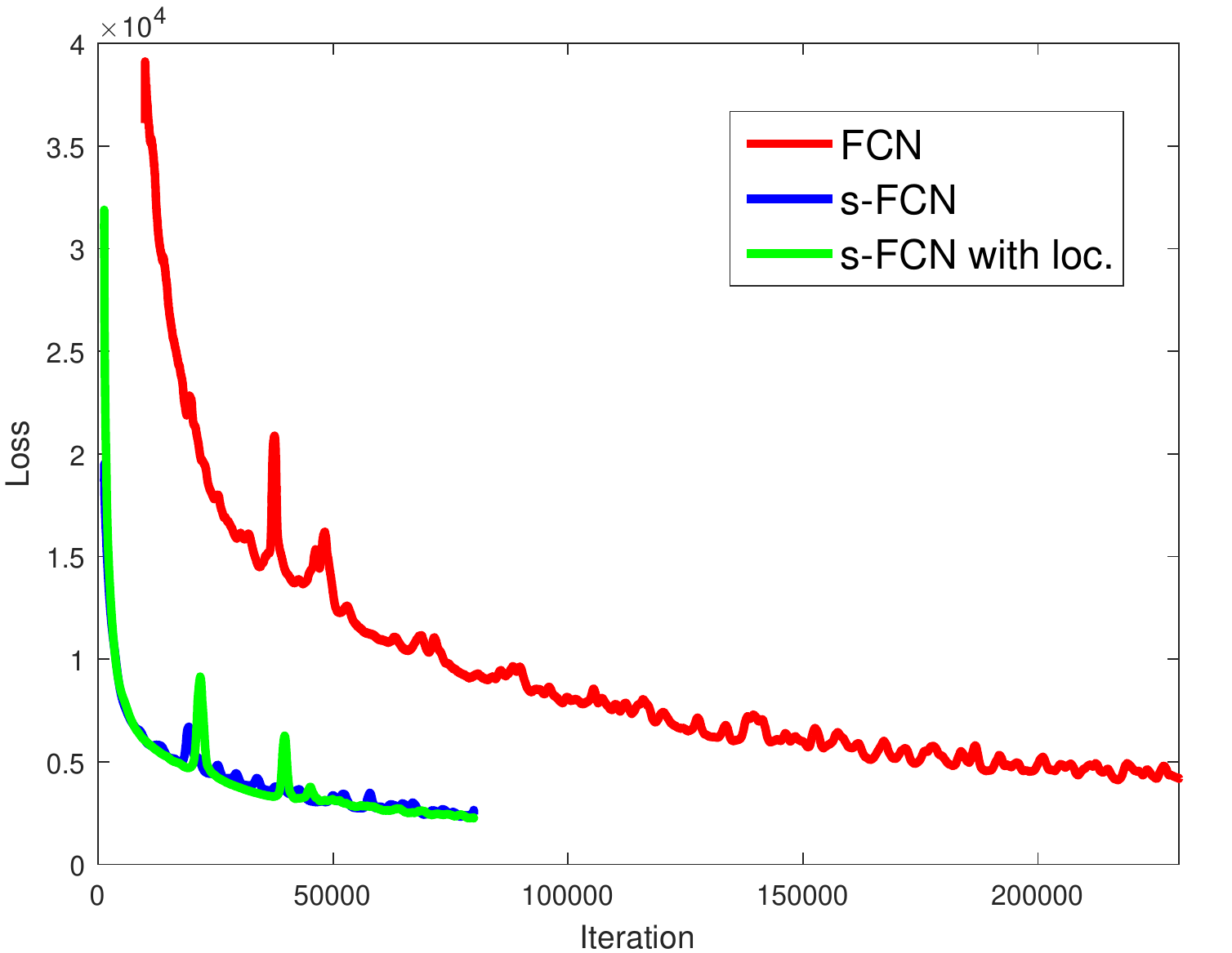} }    
		\subfigure[OC dataset.] { \label{fig_coner_2}    
			\includegraphics[width=0.46 \columnwidth]{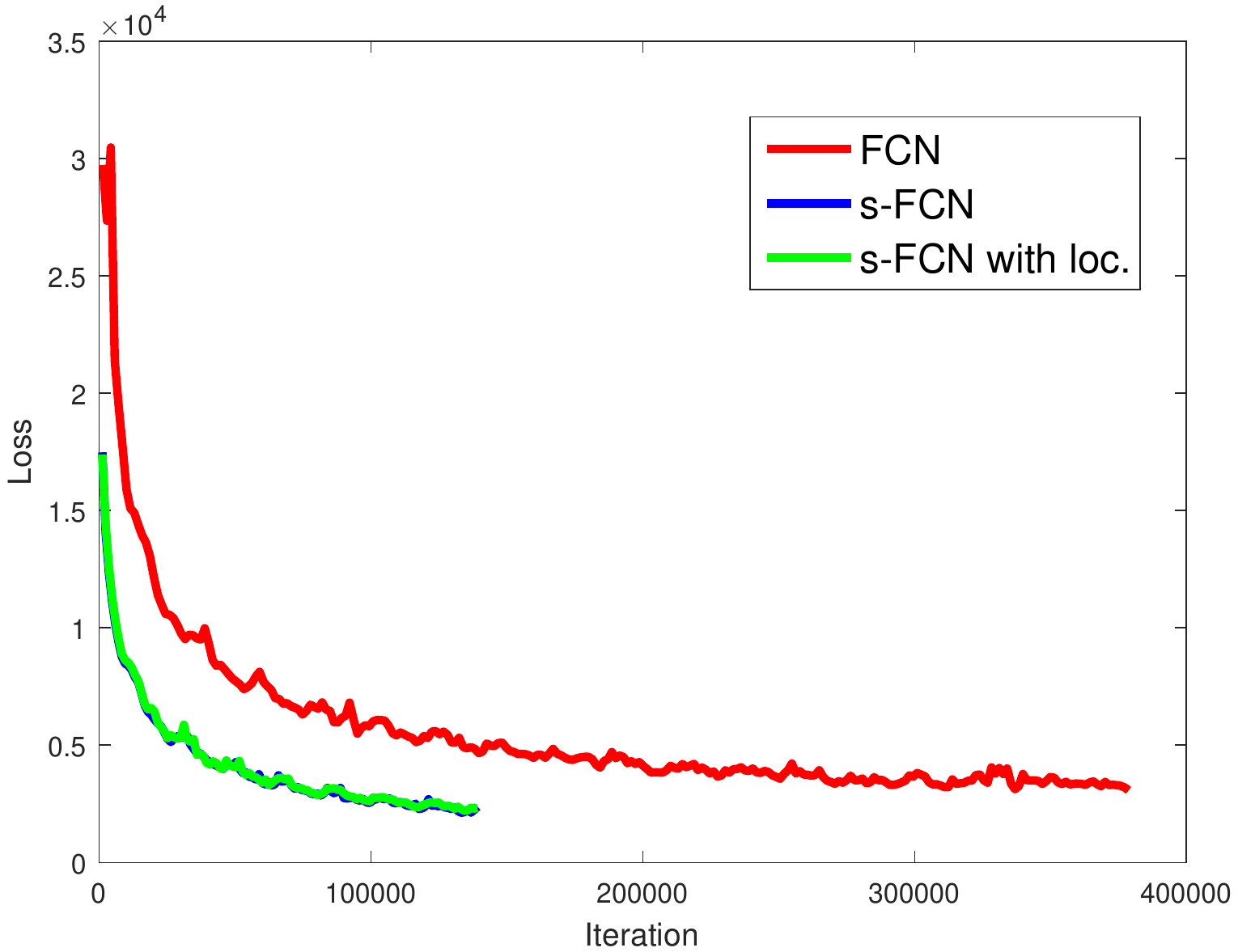} }  
		\caption{The convergent trends of FCN (red), our proposed s-FCN (blue) and s-FCN-loc (green) during the training process. (a): KITTI dataset; (b): OC dataset. }\label{Fig-conver}
	\end{figure}

	Fig. \ref{Fig-conver} illustrates the trends of convergence for three different models on the two datasets. We find the convergent speeds of s-FCN and s-FCN-loc are faster than the original FCN, and the curve lines of s-FCN and s-FCN-loc are very close during the training process. As for the KITTI dataset, the original FCN converges after $240,000$ iterations, but the proposed s-FCN and s-FCN-loc only need $80,000$ iterations to converge. In terms of iteration number, the convergent speeds of the latter two are about $70$\% faster than that of the original FCN. As a matter of fact, it is unfair to measure the convergent speed of each model by iteration number, because the computation time of each iteration is not equable for different models. Since the original FCN has only one stream, the time of its one iteration is only half of that of s-FCN and s-FCN-loc. Even so, the convergent speeds of s-FCN and s-FCN-loc are still 30\% faster than the original FCN according to the overall training time. It's worth mentioning that the difference of convergent speeds is more larger in the initial stage (the first 1,000 iterations) of training. Similarly, the consistent phenomena also appear on the OC dataset.
	
	The above phenomenon demonstrates that effective spatial structures and contour information speed up the training models. In essence, the semantic contour maps are regarded as higher-level feature than raw RGB images. The neural network can easily learn more effective semantic representation from highly structured contour maps, which guides the model to convergence more quickly. So it saves more training time than traditional single-stream network. Nevertheless, the neural network can not extract features from only semantic contour maps because it loses a lot of detailed and colorful information. Thus, it needs to have two streams to handle RGB images and semantic contour maps, respectively.

	\subsection{Comparison of the Different Location Feature Maps}
	\label{loc-feature-map}
	
	When incorporating location priors in s-FCN, two sizes of feature maps are candidates, namely $16 \times 16$ and $44 \times 44$ location maps. In Fig. \ref{Fig-featuremap}, we can find that the $44 \times 44$ feature map has more accurate location priors than the $16 \times 16$ feature map. For an input image ($500 \times 500$), the more large-size location map has the smaller  respective filed so that the location map can describe more diverse and irregular shape. Thus, the location of the distant road
	and the small corner road region can be fine represented in the former than the latter. And we believe that the former can achieve more accurate results than the latter. In order to confirm our conjecture, further contrast experiments (s-FCN with $16 \times 16$ and $44 \times 44$ location maps) are conducted on the two dataset. Table \ref{priors_table} shows the quantitative results of the two sizes of feature maps in the s-FCN-loc. From the results, the s-FCN with $44 \times 44$ feature map can achieve a higher F1-measure than the s-FCN with $16 \times 16$ feature map. Therefore, s-FCN-loc incorporates the $44 \times 44$ location feature map for the higher performance.

	\begin{table}[htbp]
		
		\centering
		\caption{Quantitative comparison of s-FCN-loc with different location priors ($16 \times 16$ and $44 \times 44$ feature maps) on the two datasets. The bold fonts represent the best performance in the corresponding column.}
		
		\begin{tabular}{c|cccc}
			\hline
			Methods 				 &F1-measure 		&Acc.			&Pre.			&Rec. 		 		\\
			\hline
			\textbf{KITTI dataset}\\
			s-FCN-loc($16 \times 16$) & 94.76 			& 98.42			&\textbf{96.54} &93.04				\\
			s-FCN-loc($44 \times 44$) &\textbf{95.38} 	&\textbf{98.56} &94.29			&\textbf{96.48} \\
			\hline
			\textbf{OC dataset} \\
			s-FCN-loc($16 \times 16$) &97.50			&98.05			&96.80		 	&\textbf{98.22} \\
			s-FCN-loc($44 \times 44$) &\textbf{97.56} 	&\textbf{98.10} &\textbf{97.24}	&97.88  \\
			\hline
			
		\end{tabular}\label{priors_table}
	\end{table}

	\subsection{Generalization Analysis}
	\label{gen}
	Since the above datasets are too small (579, 755 images in KITTI and OC dataset, respectively) for deep learning methods, the experiments are conducted in order to demonstrate the generalization of the proposed s-FCN-loc. To be specific, the training dataset is a combination of CityScapes's 2,975 \mbox{\cite{cordts2016cityscapes}} and CBCL StreetScenes's 3,547 images \mbox{\cite{bileschi2006streetscenes}}, totaling 6,522 training samples. The testing is conducted on the KITTI, SC, Cityscapes and CamVid \mbox{\cite{brostow2008segmentation}} datasets to show the generalization of the proposed methods.

	\begin{figure*}
		\centering  
		\subfigure { \label{KITTI_g} 
			\includegraphics[width=0.6 \columnwidth]{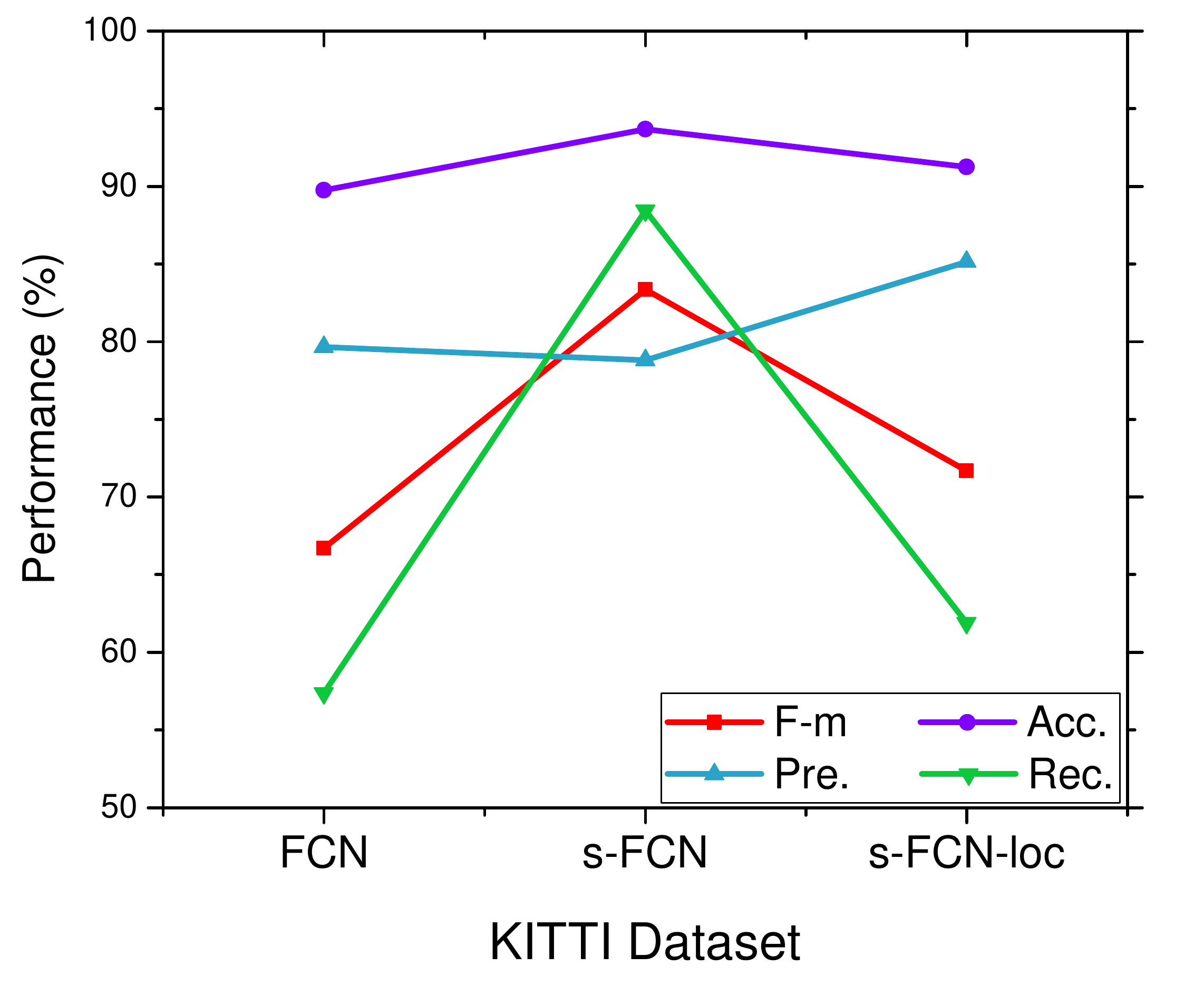} }    
		\subfigure { \label{OC_g}    
			\includegraphics[width=0.6 \columnwidth]{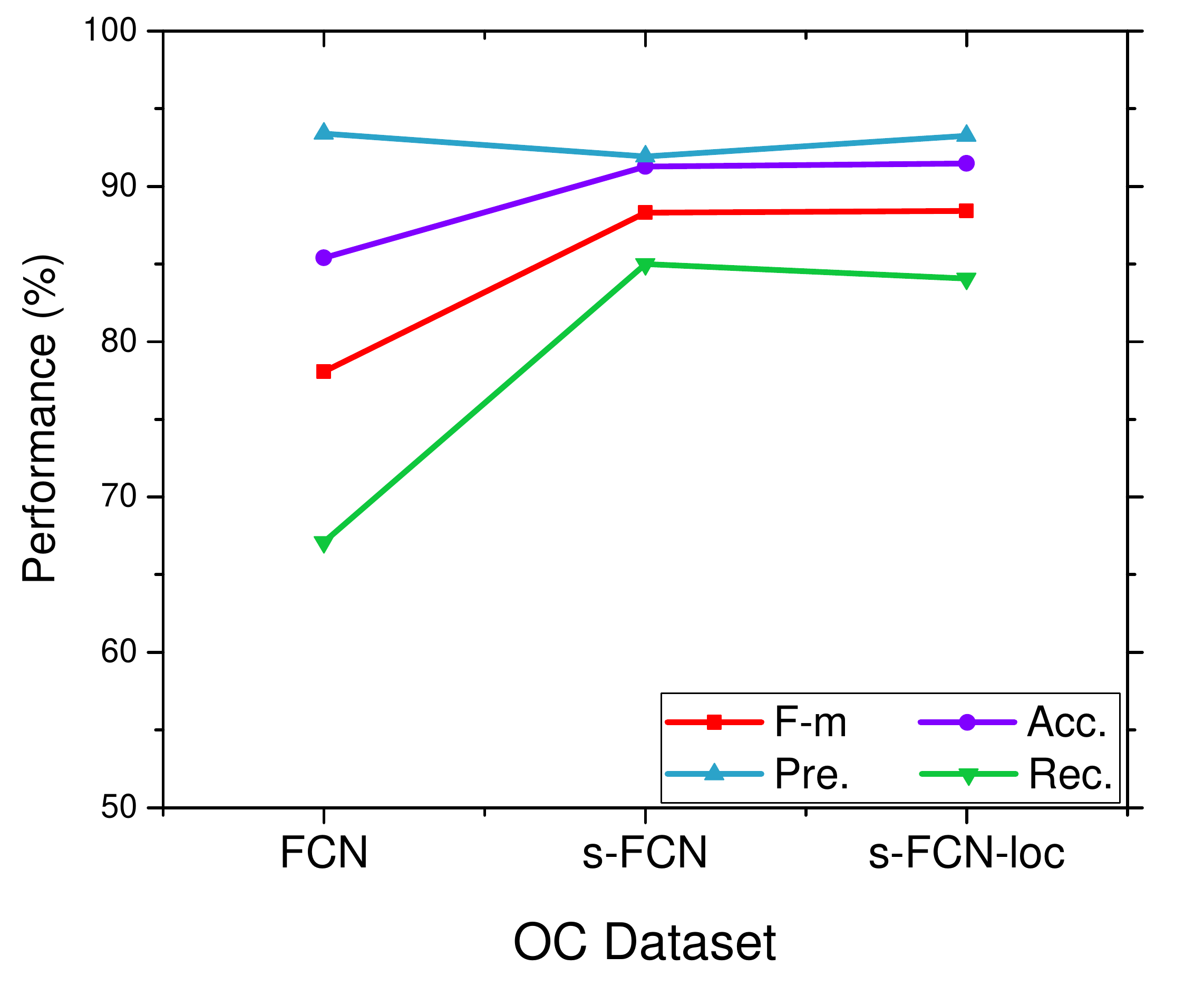} }
		\subfigure { \label{CamVid_g}    
			\includegraphics[width=0.6 \columnwidth]{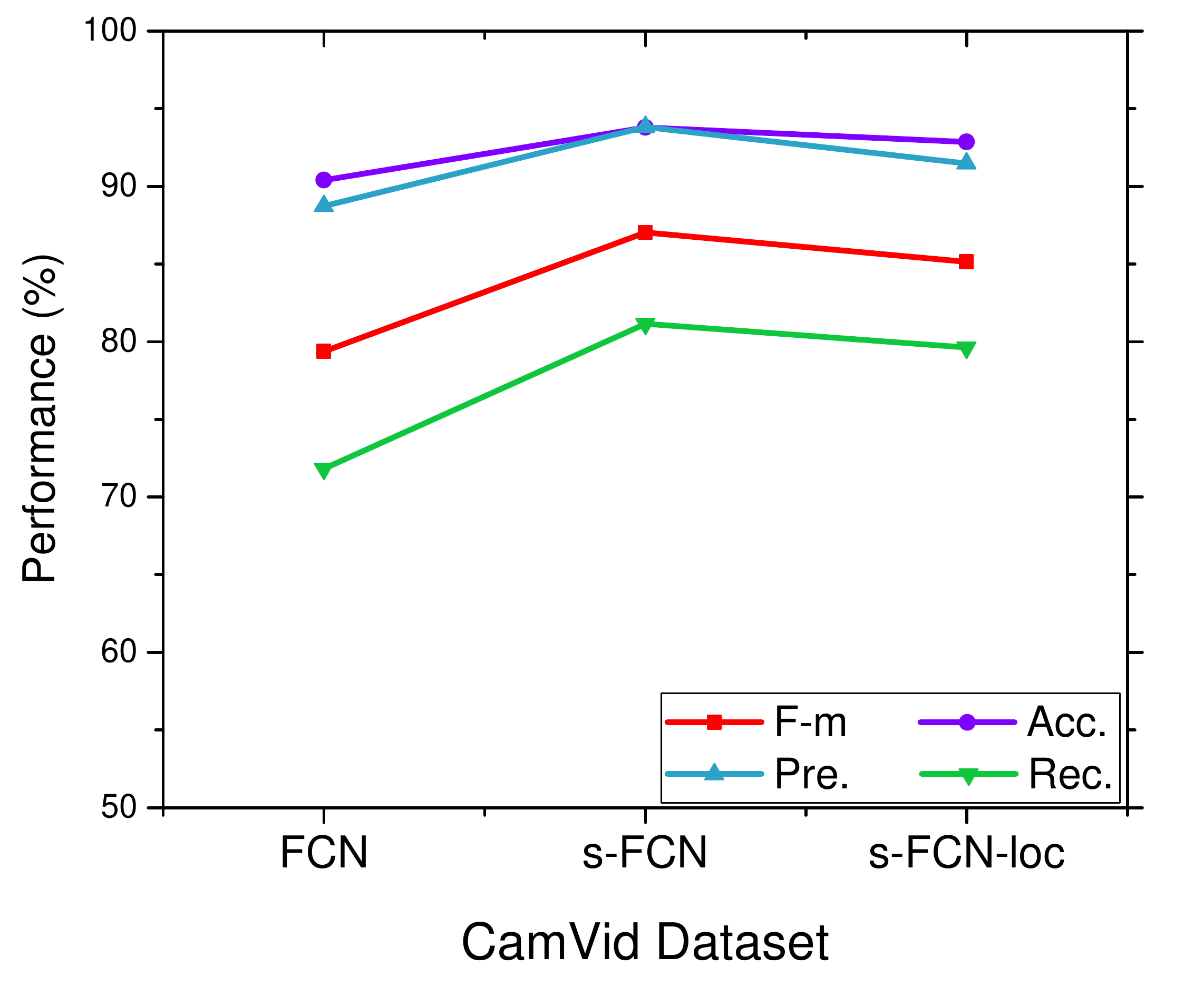} }       
		\caption{Generalization on the different testing datasets: KITTI, OC and CamVid. All models are trained on the combination of CityScapes and CBCL StreetScenes datasets.}    \label{generalization_pic}  
	\end{figure*}

	Fig. \mbox{\ref{generalization_pic}} illustrates the generalization of the three models (FCN, s-FCN and s-FCN-loc). Overall, the generalization of s-FCN and s-FCN-loc is better than that of FCN. As for the original FCN, the performance on different testing data is very poor. The essential reasons are: 1) the road has not constant shapes and structures like vehicles or pedestrians, so the network tend to learn the local appearance features (such as texture, color information); 2) the different cities adopt the different materials to build roads, and the different grades of the road need different materials, which cause the appearance features are distinct. For the proposed s-FCN and s-FCN-loc, both of them consider the higher structured information of the global scene, which is scarcely affected by changes of scenes. And the network can learn the robust structured features to represent the objects, which is an important complement. Thus, the generalization abilities of them is stronger than the original FCN.
	
\begin{figure}
		\centering
		\includegraphics[width=0.48\textwidth]{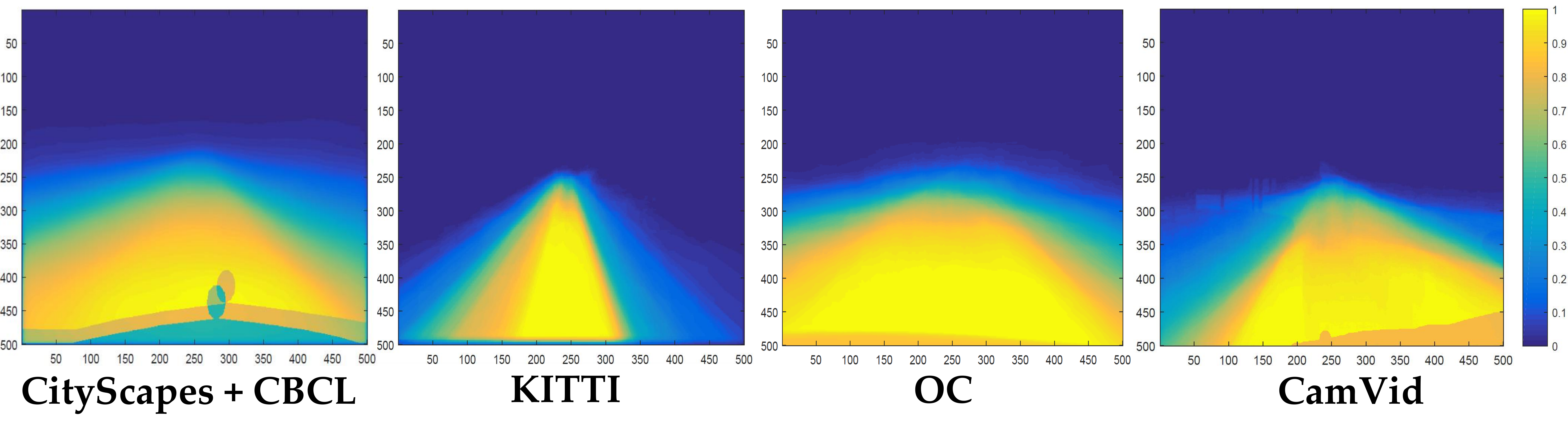}       
		\caption{The frequencies of road distribution on the four datasets: CityScapes + CBCL, KITTI, OC and CamVid. } \label{loc_p}  
	\end{figure}

	Furthermore, we find that the generalization capacity of s-FCN-loc is weaker than s-FCN on KITTI and CamVid datasets. The substantial reason is that the location priors are different because of the different camera's properties. Fig. \mbox{\ref{loc_p}} shows the frequencies of road distribution on the four datasets. The location priors of KITTI and CamVid are quite distinct from that of the combination of CityScapes and CBCL. Thus, the four metrics of s-FCN-loc are clearly inferior to that of s-FCN. On the contrary, the location prior of OC is similar to that of the combination dataset. Hence, the performance of s-FCN-loc is superior to s-FCN, which is consistent with Section \mbox{\ref{KITTI-ex1}} and \mbox{\ref{OC-ex}}. In summary, given plenty of diversified training data from the camera with the same parameters, s-FCN-loc will work better than s-FCN.
	
	\subsection{The Effects of Contour Maps vs. Depth Maps}
	\label{com}
	Generally, providing extra data sources may prompt the accuracy and convergent speed for the same model on some tasks. However, we think the contour map has instinctive advantages for road detection than other data sources: 1) the contour map is a processed, highly structured feature, which can represent the edge intensity of the global scene; 2) it can significantly improve the segmentation of the road boundaries; 3) the generation of it is easy and super real-time.

	\begin{table}[htbp]
		\centering
		\caption{Comparison of different stepwise model (the original FCN-16s, s-FCN, s-FCN with depth and s-FCN-loc ) on CityScapes validation set (in \%).}
		\begin{tabular}{c|cccc}
			\whline
			Methods 				 &F1-measure 		&Acc.				&Pre.			&Rec. 		 		\\
			\hline
			Baseline(FCN)		 	 &94.68			    &96.46				&93.69			&95.70			\\		
			s-FCN 	 			 	 &95.15				&96.74 				&93.37			&\textbf{97.01}	  		\\
			s-FCN with depth		 &93.47 		    &95.57 				&90.90 			&96.19			  		\\
			s-FCN-loc				 &\textbf{95.36}    &\textbf{96.92}		&\textbf{94.63}	&96.11			  		\\
			\whline		
		\end{tabular}\label{CS_stage}
	\end{table}
	
	\begin{figure}
		\centering
		\includegraphics[width=0.4\textwidth]{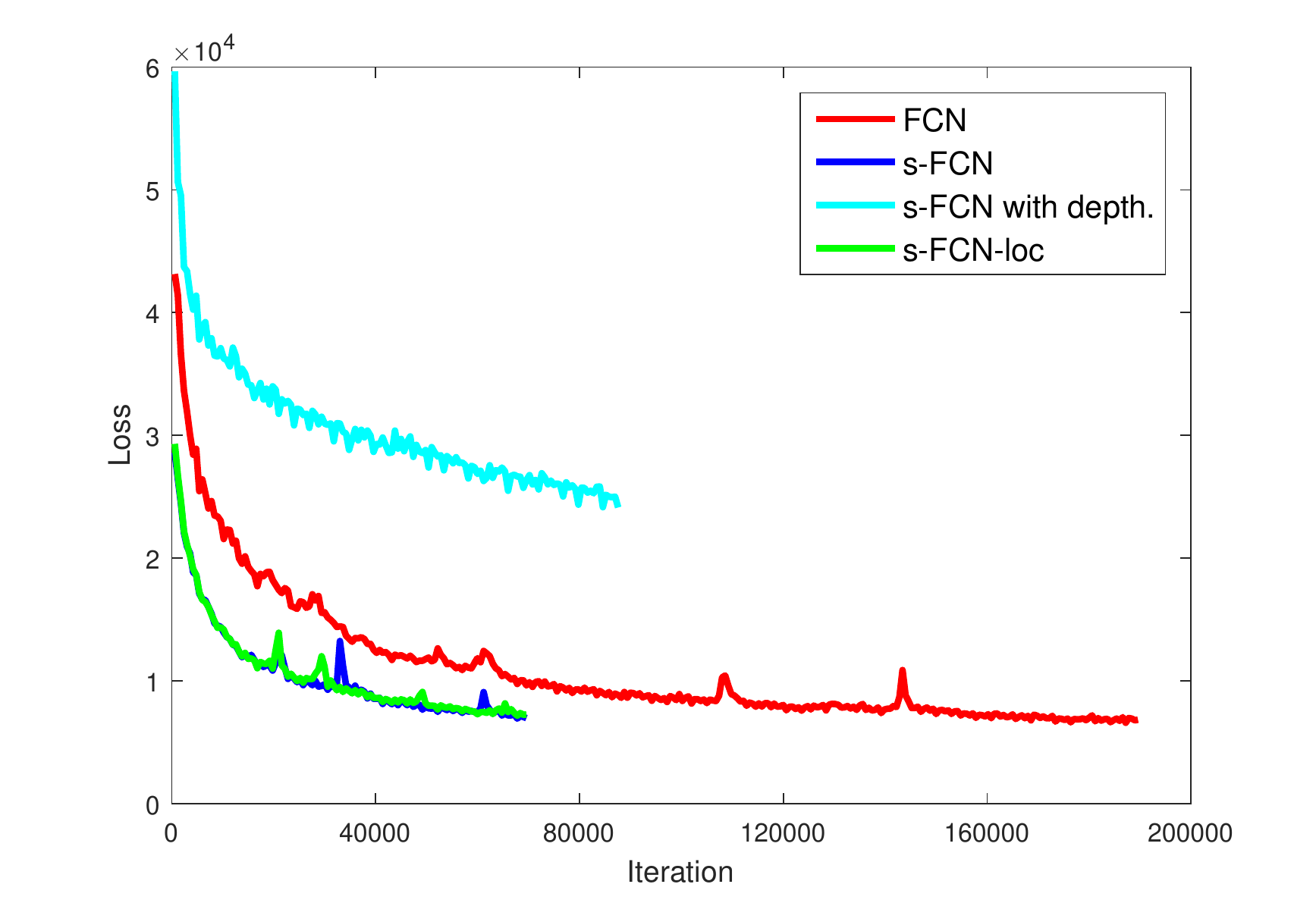}   
		\caption{The convergent trends of FCN (red), s-FCN (blue) s-FCN with depth (cyan) and s-FCN-loc (green) during the training process on CityScapes Dataset. }\label{Fig-CS_loss}
	\end{figure}
	
	Furthermore, we compare the effects of contour map and depth information on CityScapes Dataset. CityScapes is a unique large dataset that contains 2,975 training samples, and provides the RGB data as well as the depth information. The s-FCN's contour input is replaced with depth input, which is called as ``s-FCN with depth''. Quantitative results are listed in Table \mbox{\ref{CS_stage}}. In terms of the four metrics, s-FCN thoroughly defeats the s-FCN with depth. In addition to accuracies, the training time of s-FCN is less than that of s-FCN with depth from Fig. \mbox{\ref{Fig-CS_loss}}. To be specific, the learning rate is ${\rm{1}}{{\rm{0}}^{{\rm{ - 12}}}}$ during training s-FCN with depth and ${\rm{1}}{{\rm{0}}^{{\rm{ - 10}}}}$ in other models. When given learning rate of ${\rm{1}}{{\rm{0}}^{{\rm{ - 10}}}}$, the s-FCN with depth model cannot converge. Actually, since the depth maps are raw data, the model is hard to quickly learn the effect representation from it. On the contrary, contour maps have three above-mentioned advantages to improve the original FCN and speed training up.

	\section{CONCLUSIONS}
	\label{conclusion}
	This paper presents an s-FCN-loc model based on VGG-net for road detection, which is able to learn discriminative features of road boundaries and location priors. Specifically, the RGB-channel image, the semantic contour and the location prior are simultaneously integrated into a neural network without any postprocessing. Stepwise experimental results verify the effectiveness of each component in the proposed method. We also find that the proposed s-FCN-loc converges faster than the original FCN during the training stage, which saves more training time. 
	
	In the proposed s-FCN-loc, the contour stream can also be added to other networks (for instance, CNN and DeConv NN) to promote the capacity. Thus, we will transform the thought to other dense prediction tasks, such as saliency detection and semantic image segmentation in the future.
	

	\bibliographystyle{IEEEtran}
	\bibliography{IEEEabrv,reference}

	\begin{IEEEbiography}[{\includegraphics[width=1in,height=1.25in,clip,keepaspectratio]{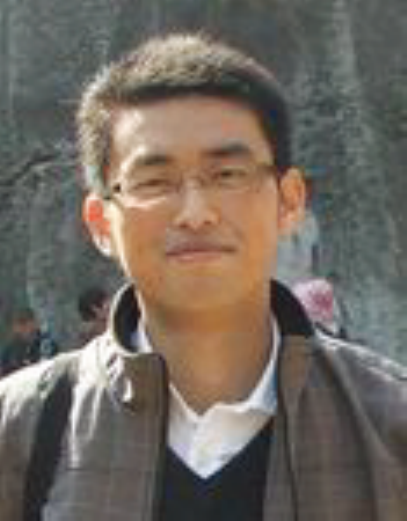}}]{Qi Wang} (M'15-SM'15) received the B.E. degree in automation and Ph.D. degree in pattern recognition and intelligent system from the University of Science and Technology of China, Hefei, China, in 2005 and 2010 respectively. He is currently a Professor with the School of Computer Science, with the Unmanned System Research Institute, and with the Center for OPTical IMagery Analysis and Learning, Northwestern Polytechnical University, Xi'an, China. His research interests include computer vision and pattern recognition.
	\end{IEEEbiography}

	\begin{IEEEbiography}[{\includegraphics[width=1in,height=1.25in,clip,keepaspectratio]{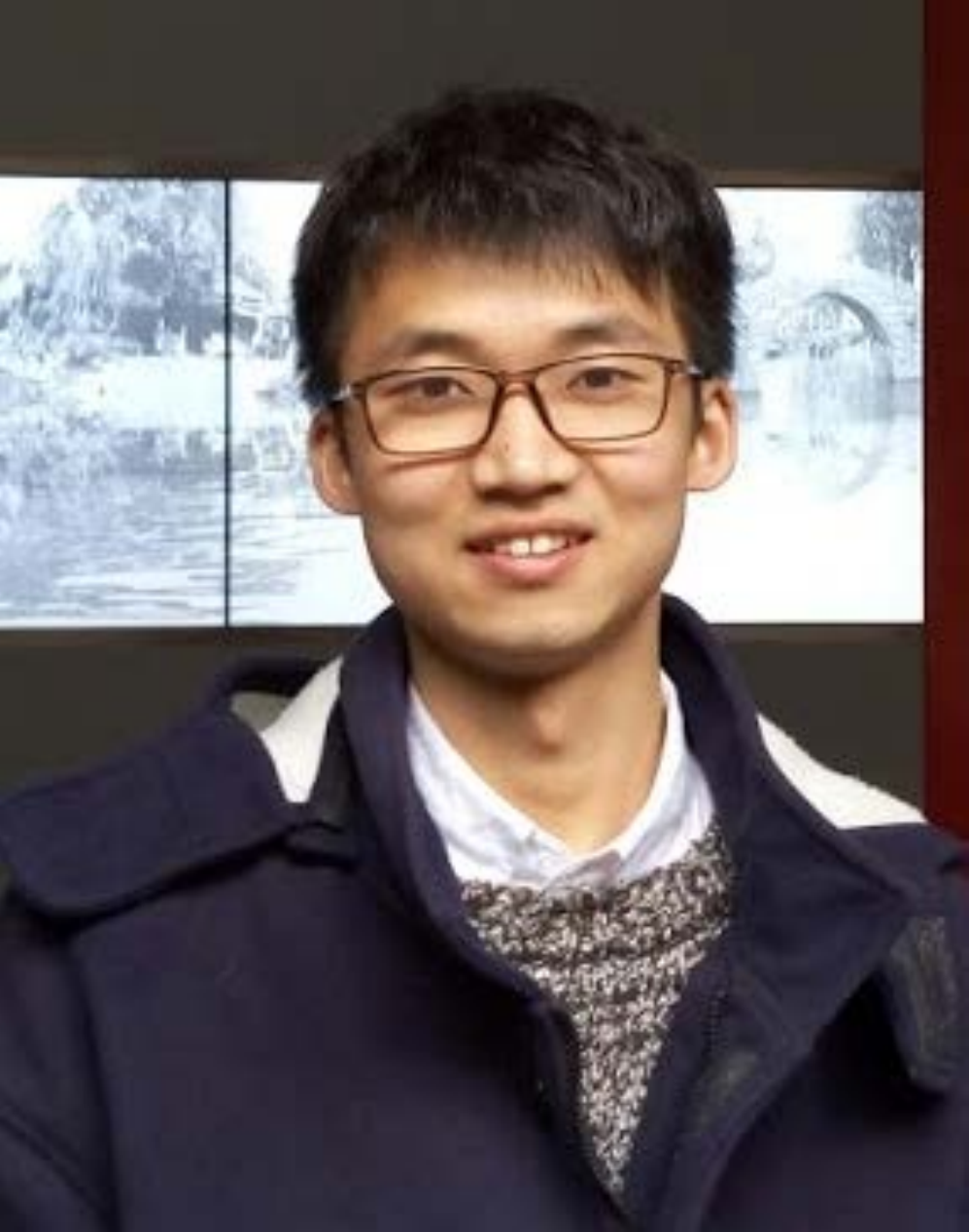}}]{Junyu Gao} received the B.E. degree in computer science and technology from the Northwestern Polytechnical University, Xi'an 710072, Shaanxi, P. R. China, in 2015. He is currently pursuing the Master degree from Center for Optical Imagery Analysis and Learning, Northwestern Polytechnical University, Xi’an, China. His research interests include computer vision and pattern recognition.
	\end{IEEEbiography}
	
	\begin{IEEEbiography}[{\includegraphics[width=1in,height=1.25in,clip,keepaspectratio]{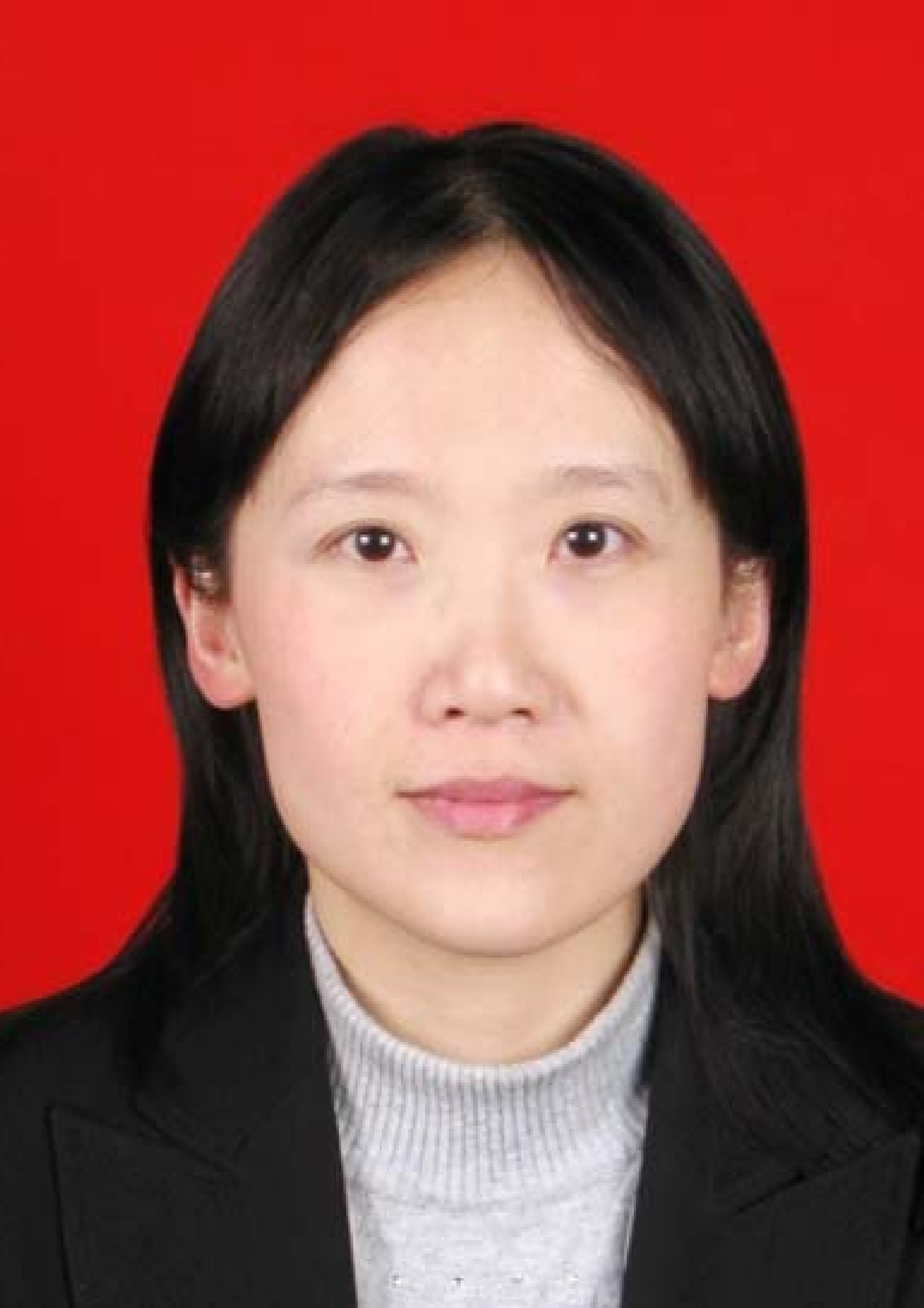}}]{Yuan Yuan} (M'05-SM'09) is currently a full professor with the School of Computer Science and Center for OPTical IMagery Analysis and Learning (OPTIMAL), Northwestern Polytechnical University, Xi'an 710072, Shaanxi, P. R. China. She has authored or coauthored over 150 papers, including about 100 in reputable journals such as IEEE Transactions and Pattern Recognition, as well as conference papers in CVPR, BMVC, ICIP, and ICASSP. Her current research interests include visual information processing and image/video content analysis.
	\end{IEEEbiography}

\end{document}